\documentclass[lettersize,journal]{IEEEtran}
\usepackage{makecell}
\usepackage{amsmath,amsfonts}
\usepackage{dsfont}
\usepackage{algorithm}
\usepackage{algorithmic}
\usepackage{hyperref}
\usepackage{cleveref}
\usepackage{array}
\usepackage{courier}
\hypersetup{hidelinks}
\usepackage{threeparttable} 
\usepackage{subfigure}
\usepackage{textcomp,soul}
\usepackage{stfloats}
\usepackage{url}
\usepackage{verbatim}
\usepackage{graphicx}
\usepackage{booktabs}
\usepackage{multirow}
\usepackage{cite}
\usepackage{amsthm}
\usepackage{bm}
\usepackage{amssymb}
\usepackage{xspace}
\newcommand{\etc}{\textit{etc.}\xspace}
\newcommand{\eg}{\textit{e.g.}\xspace}
\newcommand{\ie}{\textit{i.e.}\xspace}

\newcolumntype{C}[1]{>{\centering\arraybackslash}p{#1}}   
\theoremstyle{definition}
\newtheorem{remark}{Remark}

\DeclareMathOperator*{\argmax}{arg\,max}
\usepackage[table]{xcolor}
\usepackage{etoolbox}
\definecolor{lightblue}{rgb}{0.7,0.9,1}
\definecolor{lightgreen}{rgb}{0.7,1,0.7}
\definecolor{newlightgreen}{RGB}{177, 207, 135}
\definecolor{mediumgray}{rgb}{0.55, 0.55, 0.55}
\definecolor{lightgray}{rgb}{0.75, 0.75, 0.75}
\definecolor{lightsalmon}{RGB}{248, 214, 190}
\definecolor{lightpurple}{RGB}{200, 176, 200}
\definecolor{newred}{RGB}{215, 43, 44}
\definecolor{fancyorange}{RGB}{242, 148, 64}
\newcommand{\second}[1]{\colorbox[HTML]{fff5b3}{#1}}
\definecolor{fancygreen}{RGB}{194, 230, 140}
\definecolor{fancyblue}{RGB}{175, 184, 255}
\definecolor{lightsalom}{RGB}{247, 228, 216}
\makeatletter
\pretocmd\@bibitem{\color{black}\csname keycolor#1\endcsname}{}{\fail}
\newcommand\citecolor[1]{\@namedef{keycolor#1}{\color{blue}}}
\makeatother

\title{PRIMA: Preference-Informed Multimodal Trajectory Planning with Vision-Language Model for Autonomous Driving}
\author{
  Ruoyu Yao, Yubin Wang, Haichao Liu, Rui Yang, Zengqi Peng, Lei Zhu, and Jun Ma, \textit{Senior Member, IEEE}
  \thanks{Ruoyu Yao, Yubin Wang, Haichao Liu, Rui Yang,  Zengqi Peng, Lei Zhu, and Jun Ma are with the Robotics and Autonomous Systems Thrust, The Hong Kong University of Science and Technology (Guangzhou), China (e-mail: 
  jun.ma@ust.hk).}
  }

\begin{document}
\maketitle
\begin{abstract}
Decision-making and motion planning constitute critical components for ensuring the safety and efficiency of autonomous vehicles (AVs). Existing methodologies typically adopt two paradigms: decision then planning or generation then scoring. However, the former architecture often suffers from decision-planning misalignment that incurs risky situations. Meanwhile, the latter struggles to balance short-term operational metrics (\eg, immediate motion smoothness) with long-term tactical goals (\eg, route efficiency), resulting in myopic or overly conservative behaviors. 
To address these issues, we introduce PRIMA, a preference-informed multimodal trajectory planning framework assisted by a Vision-Language Model (VLM). 
Our approach employs a tri-stage cognitive decomposition protocol, encompassing scene understanding, action selection, and confidence assessment, to generate diverse candidate actions paired with verbalized confidence scores. 
This reasoning mechanism exposes alternative tactical choices and provides an empirical preference signal for balancing them with downstream planning quality. 
Furthermore, we propose a novel planning module that integrates a diffusion model for trajectory generation and a hierarchical refinement process to find the optimal trajectory. 
This enables the selection over trajectory candidates accounting for both low-level solution quality and high-level tactical preferences, avoiding the risks within one-shot decisions and overcoming the limitations of short-sighted scoring mechanisms. 
Comprehensive simulation-based closed-loop evaluations on nuPlan and InterPlan demonstrate competitive performance across common and long-tail benchmarks. A paired controlled intervention further evaluates the incremental decision utility of the verbalized confidence signal. The code will be released upon acceptance.
\end{abstract}


\section{Introduction} \label{sec:intro}
\begin{figure}[t] 
\centering 
\includegraphics[width=1\linewidth]{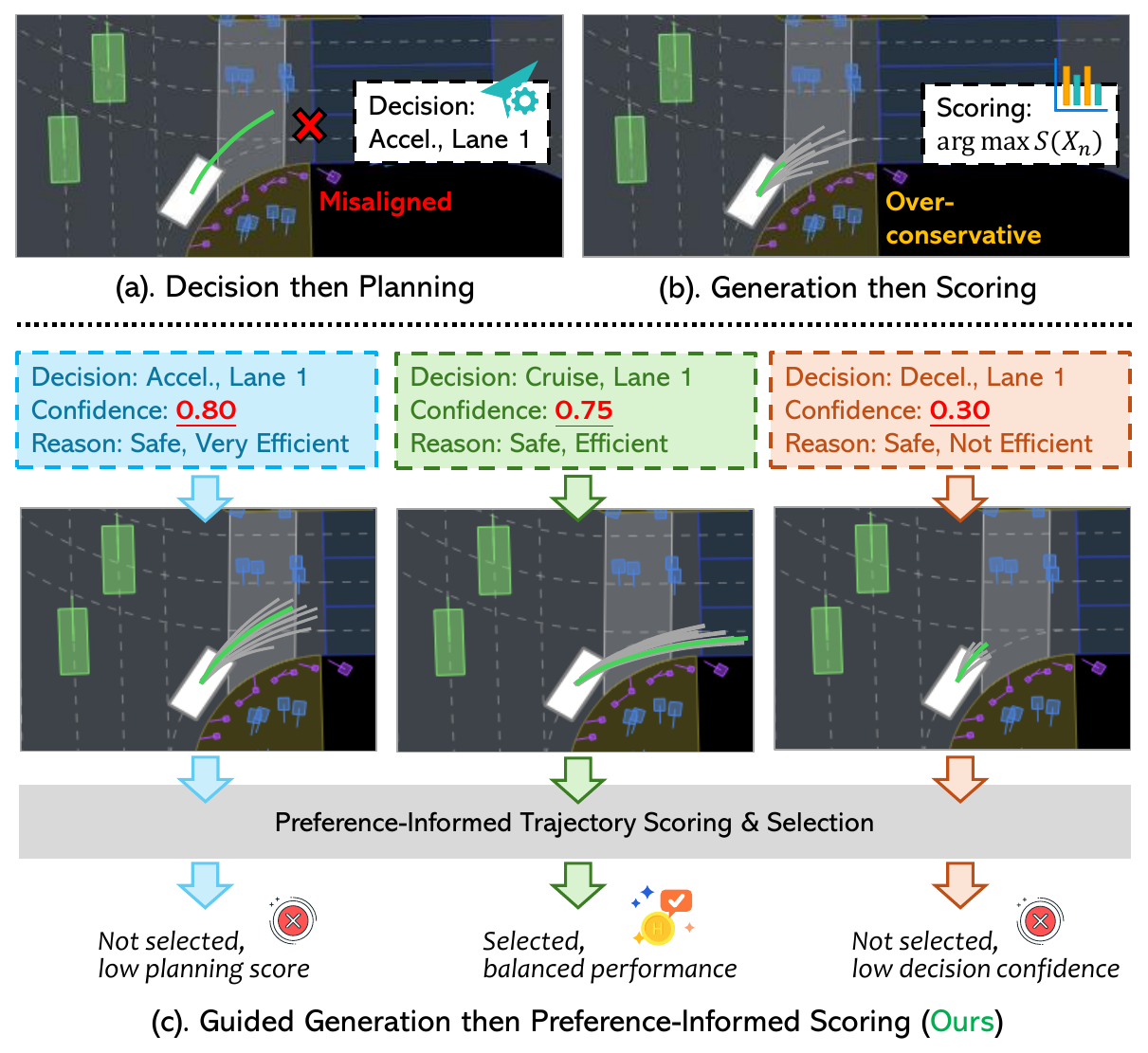} 
\caption{An illustration of decision-making and motion planning paradigms. In the shown instance: (a). The downstream planner fails to produce a trajectory that closely aligns with the high-level decision of accelerating to the rightest lane. The trajectory deviation poses a collision risk with the pedestrians. (b). The scoring process prioritizes an overly conservative trajectory that fails to account for the long-term efficiency beyond the planning horizon (\ie, avoiding waiting for all the pedestrians to pass). (c). Our approach explicitly incorporates multimodal decisions with confidence scores, enabling decision-guided trajectory generations and preference-informed scoring and selection, which allows finding the best plan balancing decision-level preferences and motion planning quality.}
\vspace{-0.5em}
\label{fig:parad compare}
\end{figure}

Decision-making and motion planning are critical components for Autonomous Vehicles (AVs), which directly influence safety, efficiency, and consistency in driving performance. Existing methodologies in this domain typically fall into two paradigms: \textit{decision then planning}~\cite{li2021safe, sha2023languagempc, yao2024hierarchical, peng2025bilevel} or \textit{generation then scoring}~\cite{zeng2019end, adajania2022multi, huang2023conditional, huang2024gen}. While the first paradigm is conceptually straightforward, it struggles to address the potential misalignment between decision-making and planning processes~\cite{zhang2021unified}. 
Recent studies show that even state-of-the-art planners could fail to closely align with high-level effective decisions made by humans, thereby leading to unsatisfactory performance~\cite{yang2024diffusion}. 
In contrast, the second paradigm generates multimodal trajectory candidates, enabling planning-informed decision-making based on explicit performance scores associated with different trajectories~\cite{cui2021lookout}. However, a challenge persists in designing an adequate scoring mechanism that facilitates the selection of the best plan in complex and dynamic real-world environments. This requires a holistic evaluation across different trajectory alternatives to ensure reasonable long-term decision-making (\eg, determining the correct timing for overtaking to gain travel efficiency)~\cite{michon1985critical}. Meanwhile, it also requires incorporating low-level operational metrics to improve the short-term planning performance (\eg, steering properly to allow for the smoothness)~\cite{werling2010optimal}.

Current research on scoring and selection processes for multimodal trajectory candidates encompasses rule-based approaches~\cite{adajania2022multi, dauner2023parting} and learning-based methods~\cite{zeng2019end, hu2022st}. Rule-based systems, integrating diverse metrics within hand-crafted functions, quantify the trajectory performance from different perspectives in an interpretable manner. These methods can exhibit excellent driving performance in common and well-structured traffic scenarios~\cite{dauner2023parting}. However, due to the fixed scoring designs, the driving systems encounter critical situations where the hand-crafted functions fail to generalize~\cite{cheng2024rethinking}. In particular, the rule-based scorers fall short of accurately reflecting the significance of long-term decision-making. This makes the systems struggle to recognize driving strategies such as detouring around road construction by temporarily occupying the opposite lane~\cite{tiandrivevlm}, unless such specific scenarios are explicitly handled by the ad hoc decision logic. On the contrary, learning-based methods are developed to automatically capture a reward or cost model for scoring trajectories through training on the data of human driving records. Representative studies in this category primarily utilize inverse reinforcement learning~\cite{rosbach2019driving, huang2023conditional} or end-to-end learning~\cite{zeng2019end, cui2021lookout}, demonstrating improved decision-making flexibility. Nevertheless, in these approaches, the training objective is  confined to facilitate domain-specific model fitting. The trained models do not gain the capacity to explicitly interpret the rationale behind the decision-making process or calibrated scoring weights. Therefore, they are not guaranteed to generalize to long-tail challenging scenarios and even make unexpected mistakes in simple situations~\cite{shao2024lmdrive, jiang2024senna}.

Recent advancements in large foundation models prompt investigations into leveraging their knowledge and reasoning capabilities to assist in decision-making~\cite{cui2024survey}. Built upon Large Language Models (LLMs) or Vision-Language Models (VLMs), the decision-making components can perform Chain-of-Thought (CoT) reasoning to derive an answer while utilizing knowledge gained from extensive pre-training tasks~\cite{shao2024lmdrive}, thereby improving generalization ability compared to traditional learning-based methods. Under the scoring-based decision-making framework, in~\cite{wang2024he}, a rule-based trajectory scorer is coupled with a VLM module, which allows the VLM to adjust the scorer weights in real-time such that the trajectories conforming to favorable driving styles can be prioritized. 
More recently, MindDrive~\cite{sun2025minddrive} integrates scene, trajectory, and language representations within a VLM-based evaluator, using a dedicated score token and scoring head to produce multi-objective evaluations of candidate trajectories. 
Despite these advances, the current integration of semantic reasoning into trajectory scoring fails to explicitly represent and compare competing high-level tactical alternatives. 
This motivates a planning interface that retains preference over diverse decision candidates and postpones the final arbitration until their associated trajectories have been evaluated.

Motivated by this, we introduce \textbf{PRIMA}, a \textbf{PR}eference-\textbf{I}nformed \textbf{M}ultimodal planning framework assisted by a VLM for \textbf{A}utonomous driving. This framework integrates a tri-stage cognitive decomposition protocol, encompassing scene understanding, action selection, and confidence assessment, to produce Top-K tactical candidates with verbalized confidence scores~\cite{tian2023just}. 
This structured elicitation exposes alternative tactical choices and provides a tactical preference cue for balancing them against downstream planning quality. 
To bridge tactical decision-making and motion planning, the framework adopts decision-guided trajectory generation followed by preference-informed cross-candidate selection. A diffusion-based planner generates proposals under the objectives associated with each action, and a hierarchical refinement procedure selects the final plan using trajectory quality together with the tactical preference term. 
An intuitive comparison of our paradigm with existing ones is depicted in Fig.~\ref{fig:parad compare}. 
The approach is intended to reduce premature commitment in one-shot decision-making and to complement short-horizon rule-based trajectory scoring with VLM-proposed tactical alternatives.
Our contributions are summarized as follows: 
\begin{itemize} 
\item We present a preference-informed, VLM-assisted multimodal planning framework for autonomous driving. A tri-stage cognitive decomposition protocol enables the decision-maker to reason over multiple candidates and verbalize confidence scores that express its relative tactical preferences for downstream arbitration.

\item We develop a guided trajectory generation and hierarchical refinement module to obtain a feasible, smooth, and long-term decision-aware trajectory by incorporating motion planning quality and decision-level preferences. This effectively addresses decision-planning misalignment inherent in both traditional and foundation model-empowered AVs.

\item Our approach achieves competitive performance on both common and long-tail benchmarks in the nuPlan closed-loop simulation~\cite{caesar2021nuplan}, and further validates strong cross-domain generalization on the challenging InterPlan benchmark~\cite{hallgarten2024can}. It outperforms the evaluated knowledge-driven baselines across core planning metrics and reduces severe driving errors in both non-reactive and reactive scenarios on Test14.

\end{itemize}

\section{Related Works}
\begin{figure*}
  \centering
    \includegraphics[width=1\linewidth]{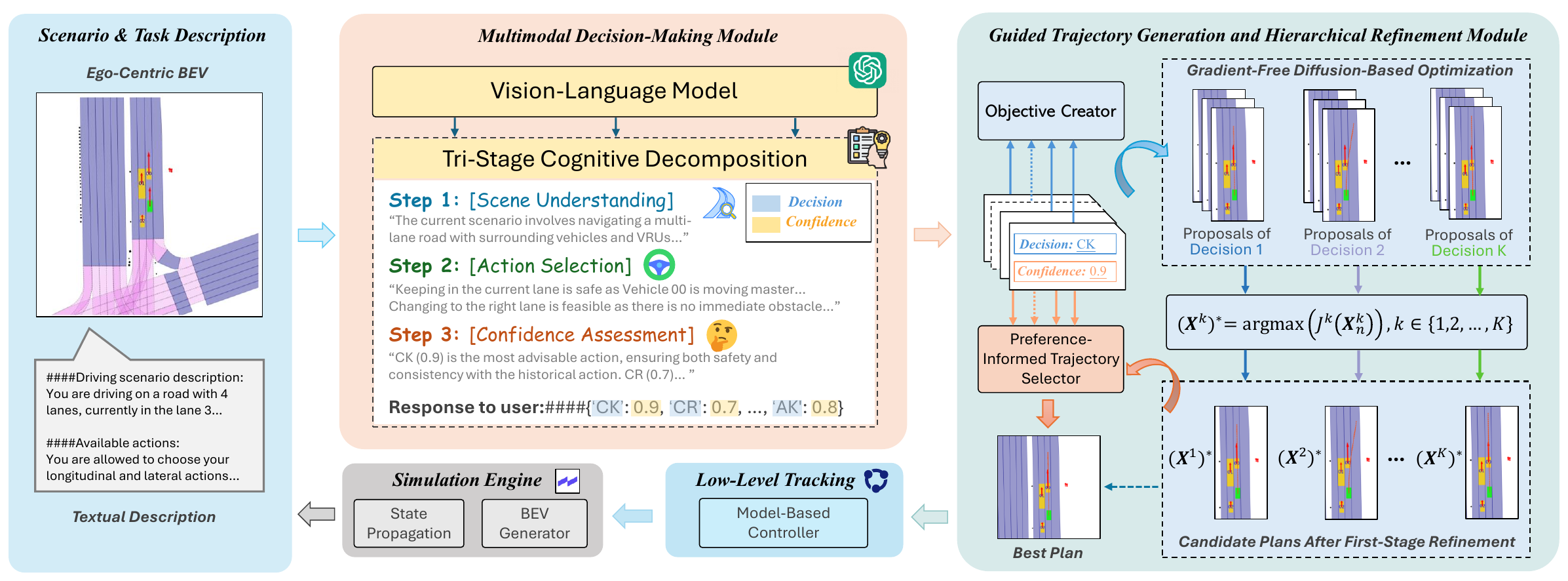}
  \caption{The PRIMA pipeline features a multimodal decision-making module based on a VLM and a motion planning module integrating decision-guided trajectory generation and hierarchical refinement. A BEV image and textual description are sent to the VLM for tri-stage cognitive reasoning to elicit Top-K candidate decisions with model-reported heuristic preference scores. These candidate decisions then guide the generation and refinement of trajectory proposals via gradient-free diffusion-based optimization. Finally, a preference-informed trajectory selector determines the best plan by combining the tactical preference signal with low-level motion-planning quality.}
  \label{fig:model_framework}
\vspace{-0.5em}
\end{figure*}

\label{sec:related work}
\subsection{Large Foundation Models for Autonomous Driving}
Large foundation models demonstrate promising common-sense reasoning capabilities for generalized AVs~\cite{cui2024survey, liu2024lmmcodrive}. Early studies leverage LLMs for textual driving context understanding, enhancing scene adaptability through CoT-based planning objective modulation~\cite{sha2023languagempc} and enabling closed-loop lifelong learning by integrating reasoning, memory, and reflection modules~\cite{wendilu}. To bridge the modality gap, recent works adopt VLMs to synthesize cross-modal representations. ComDrive employs LLaMA-3.2V for driving scenario comprehension based on multi-view images and a driving-oriented dialog context, and applies adjustments to the trajectory scoring function to ensure smooth planning~\cite{wang2024he}. PlanAgent utilizes GPT-4V to perform hierarchical CoT reasoning and generate motion planning parameters from a Bird's Eye View (BEV) image combined with textual descriptions~\cite{zheng2024planagent}. Meanwhile, notable efforts are devoted to Vision-Language-Action (VLA) integration to align semantic reasoning with end-to-end driving~\cite{xu2024drivegpt4, shao2024lmdrive, zhou2026opendrivevla}.

Despite these advancements, existing methods predominantly treat the foundation model as a deterministic reasoning engine that maps a complex scenario to a single output, leaving no explicit alternative strategies or model-reported preference signal for downstream arbitration.
Our framework departs from this single-output paradigm by eliciting multiple candidate strategies and a VLM-reported confidence score for each candidate. These candidates directly guide downstream multimodal planning and allow trajectory selection to combine the resulting relative preference signal with execution quality.

\subsection{Confidence Elicitation in Large Foundation Models} 
Uncertainty estimation has been widely studied in deep learning. 
Bayesian neural networks~\cite{neal2012bayesian, blundell2015weight}, Monte Carlo dropout~\cite{gal2016dropout}, and deep ensembles~\cite{lakshminarayanan2017simple} provide established approaches for characterizing model variability. 
However, these methods scale poorly to large foundation models like LLMs and VLMs due to the excessive cost of re‑sampling and ensembling. 
Logit-based methods offer a more efficient alternative~\cite{huang2023look}, but they generally require access to token-level probabilities and may be affected by sequence length, semantic equivalence, and token importance imbalance~\cite{geng2024survey, kuhnsemantic, duan2024shifting}. These requirements limit their applicability to closed-source foundation models. 

Confidence elicitation provides another interface for obtaining response-level confidence information from large foundation models~\cite{kadavath2022language}. Instead of estimating uncertainty through repeated inference or token probabilities, the model is prompted to verbalize its confidence in a generated answer~\cite{lin2022teaching}, without requiring access to internal logits or modification of the model architecture. Subsequent studies further develop this black-box elicitation approach from complementary perspectives. Tian et al. evaluate prompting strategies to elicit verbalized confidence from LLMs fine-tuned through Reinforcement Learning with Human Feedback (RLHF), showing that considering alternative answers can improve confidence estimates~\cite{tian2023just}. Xiong et al. formulate black-box elicitation as a combination of prompting, response sampling, and consistency-based aggregation, benchmarking the framework for confidence assessment and failure prediction~\cite{xiongcan}.

In autonomous driving, confidence elicitation retains multimodal action hypotheses, reducing premature commitment to a single model output. The elicited scores provide relative preference information over candidate actions, enabling planners to incorporate high-level tactical cues alongside low-level execution quality.

\section{Methodology}\label{sec:method}
The overall architecture of PRIMA is illustrated in Fig.~\ref{fig:model_framework}, which consists of two main components: a decision-maker built upon an VLM to perform tri-stage cognitive reasoning for driving decisions, and a trajectory generation and refinement module that takes the guidance of the VLM-based decision-maker to generate motion planning proposals and hierarchically select the desired trajectory. 
This design produces a ranked set of tactical hypotheses accounting for VLM inference stochasticity, rather than committing to a single potentially hallucinated action.

\subsection{Vision-Language Model for Embodied Decision-Making}

\subsubsection{Multimodal Context Input}
To enhance the VLM's understanding of the driving environment and task, we provide a BEV image \( \bm{I}_t \) and a textual description \( \bm{D}_t \) at each time-step \( t \). This dual-modal context captures macro-level scene attributes (\eg, road structure, traffic density), fine-grained object features, and decision-making reasoning requirements.

\begin{figure}[t]
  \centering
   \includegraphics[width=0.7\linewidth]{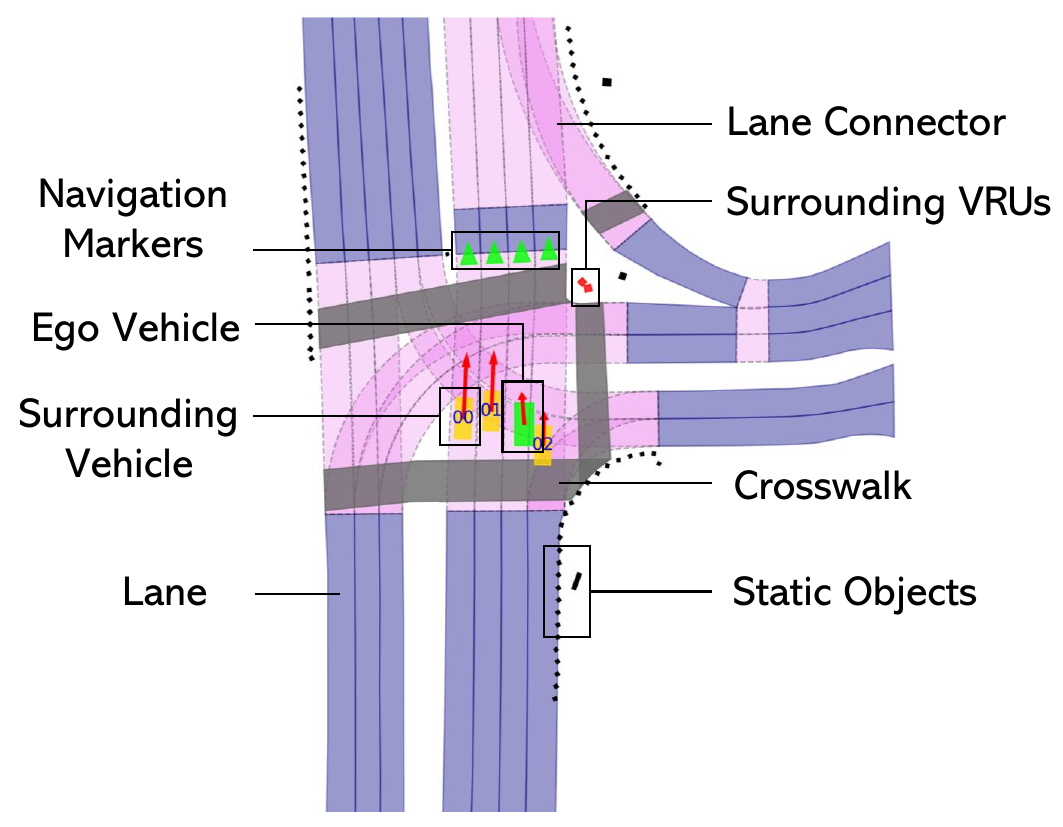}
   \caption{An illustration of different objects presented on the BEV. Annotation specifications are given by the system message.}
   \label{fig:BEV annotation}
\end{figure}

\noindent \textbf{BEV Context Representation.}
We generate a \( 100 \times 100 \, \text{m} \) ego-centric BEV image (ego facing upward) illustrating:
\begin{itemize}
    \item \textbf{Map Objects}: lanes, lane connectors, and crosswalks.
    \item \textbf{Moving Objects}: ego vehicle, surrounding vehicles, and Vulnerable Road Users (VRUs: pedestrians/cyclists), annotated with velocity arrows.
    \item \textbf{Static Objects}: all dataset-recorded static obstacles.
    \item \textbf{Navigation Markers}: junction target-lane exits based on global routing.
\end{itemize}
Objects use specific colors and shapes to convey semantics (Fig.~\ref{fig:BEV annotation}), and their interpretations are included in \( \bm{D}_t \).

\noindent \textbf{Textual Description.}
\( \bm{D}_t \) comprises a system message (outlining the task, BEV rules, and response format) and a human message (conveying navigation commands, ego/object states, ego action space/history, and decision-making guidelines). 

Following~\cite{wendilu}, we employ a scenario-based prompting strategy based on the ego-junction distance \(d_t = \inf_{\bm{x}\in\mathcal{J}}\|\bm{x}_t-\bm{x}\|_2\), where \(\bm{x}_t\) is the ego position and \(\mathcal{J}\) is the nearest junction area per global routing. With a threshold \(\bar{d} = 20 \, \text{m}\), scenarios are categorized as:
\begin{itemize}
    \item \(d_t > \bar{d}\): Normal multilane driving (no immediate junction).
    \item \(0 < d_t \le \bar{d}\): Multilane driving approaching a junction.
    \item \(d_t = 0\): Driving within a junction.
\end{itemize}

When \(d_t \le \bar{d}\), a navigation command (`go straight', `turn left', or `turn right') is added to focus the VLM on the target area. Relevant moving objects (selected according to \Cref{tab:mov obstc selec}) are described by their distance, Line-of-Sight (LoS) angle, speed, heading, and lane. 

The ego action space $\mathcal{A}$ is defined as the Cartesian product of longitudinal and lateral actions:
\begin{equation}
\begin{aligned}
    \mathcal{A} &:= \mathcal{A}_{lon} \times \mathcal{A}_{lat}, \quad \mathcal{A}_{lon} := \{a_{acc}, a_{dec}, a_{cruise}\},\\
    \mathcal{A}_{lat} &:= 
    \begin{cases} 
        \{a_{left}, a_{right}, a_{keep}\}, & \text{if } d_t > 0, \\
        \{a_{nav}\}, & \text{if } d_t = 0,
    \end{cases}
\end{aligned}
\end{equation}
where \(\mathcal{A}_{lon}\) includes accelerating, decelerating, and cruising. \(\mathcal{A}_{lat}\) comprises left/right lane-changes and lane-keeping for multilane driving, but restricts to navigation-following (\(a_{nav}\)) within junctions for safety. 

Additionally, the historical actions of the last two time-steps are provided as short-term memory for consistency. Finally, a decision-making guideline (\Cref{tab:decision_guideline}) emphasizing safety, efficiency, and comfort prompts the VLM to derive Top-K decisions through CoT reasoning across three steps: scene understanding, action selection, and confidence assessment.

\begin{table}[t]
\caption{Moving objects selection for different driving scenarios. N-MLD: Normal Multilane Driving. MLD-WAJ: Multilane Driving While Approach a Junction. JD: Junction Driving.}
\centering
\renewcommand{\arraystretch}{1.5}
\resizebox{1\linewidth}{!}{
\begin{tabular}{c|c|C{1.4cm}C{1.4cm}C{1.4cm}}
\hline
\textbf{Object Type} & \textbf{Spatial Relation} & \textbf{N-MLD} & \textbf{MLD-WAJ} & \textbf{JD} \\ \hline
\multirow{4}{*}{Vehicle} & In the Same Lane & \checkmark & \checkmark & - \\
 & In Adjacent Lanes & \checkmark & \checkmark & - \\ 
 & At the Junction & - & \checkmark & \checkmark \\
 & On the Target Road & - & - & \checkmark \\ \hline
\multirow{1}{*}{VRU} & In the Field of View\footnotemark & \checkmark & \checkmark & \checkmark \\ \hline
\end{tabular}
}
\label{tab:mov obstc selec}
\end{table}

\footnotetext{Specifically, we define the ego vehicle's field of view as a sector region centered at its position, with a radius of $30 \, \text{m}$ and a line-of-sight angle spanning from $-75^{\circ}$ to $+75^{\circ}$ relative to the vehicle's forward direction.}

\subsubsection{Tri-Stage Cognitive Reasoning for Driving Decisions}
\begin{figure}[t]
  \centering
   \includegraphics[width=0.95\linewidth]{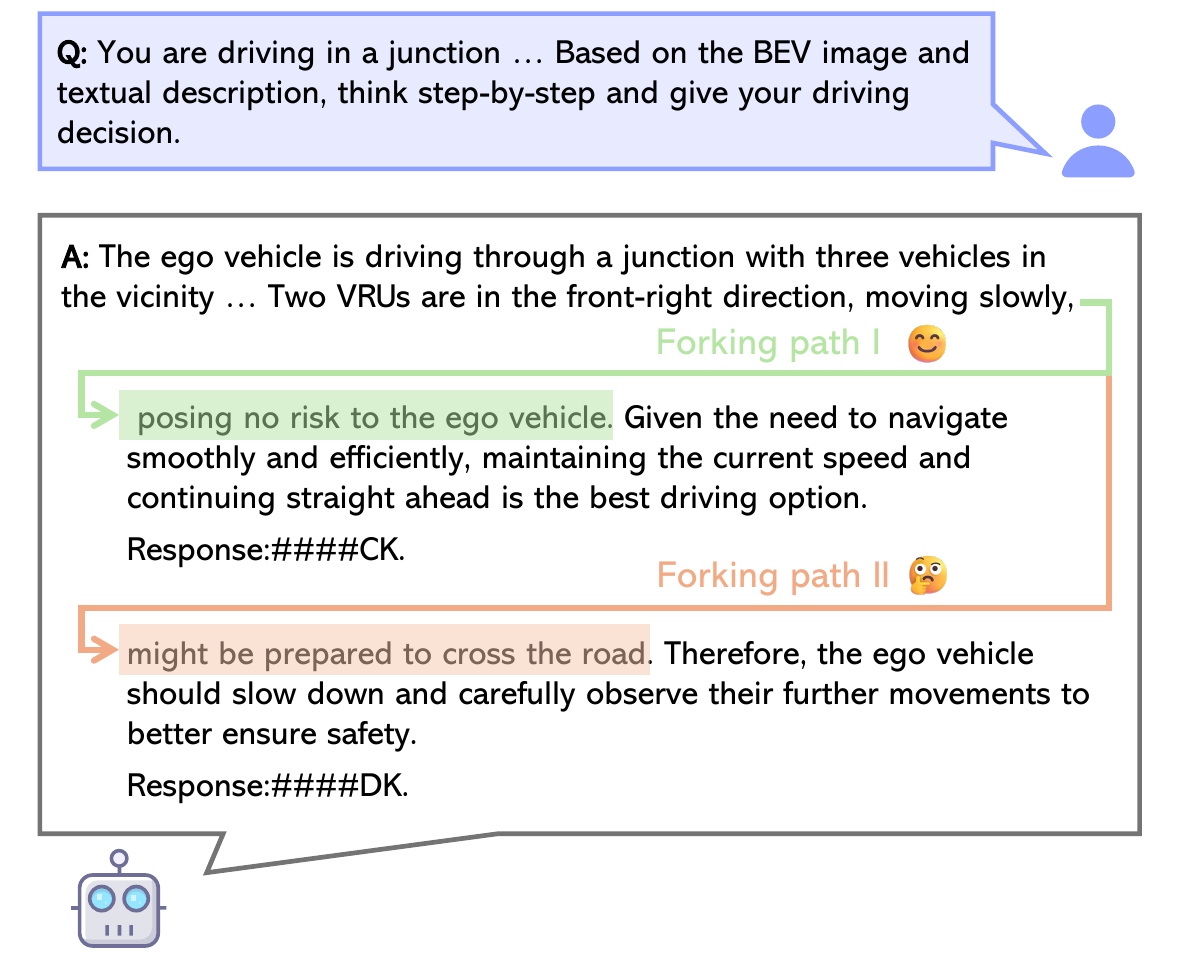}
   \caption{An example of forking paths in the generation process of VLMs. Owing to the stochasticity in an intermediate reasoning step, the final response made by the VLM can be different in multiple dialogs with the same question.}
   \label{fig:forking path}
\vspace{-0.5em}
\end{figure}

Driving involves decision-making processes with uncertainty. Given the same driving context, multiple types of decisions (\eg, accelerate and overtake the leading vehicle or cruise and follow the leading vehicle) could be deemed feasible. Previous large foundation model-empowered AVs were designed to map the driving context to a single decision~\cite{wendilu, zheng2024planagent}, which cannot represent the stochastic decision-making nature. Moreover, due to the auto-regressive generation and CoT reasoning strategy, the foundation model's final answer conveyed by the last few tokens is largely determined by the previous tokens. This means that variations occurring during the model's intermediate generation can cascade into significantly different answers, known as \textit{forking paths} in neural text generation~\cite{bigelowforking}. An example of forking paths in the question-answering of driving decision-making is shown in Fig.~\ref{fig:forking path}, with the corresponding driving scenario illustrated in Fig.~\ref{fig:BEV annotation}. Therefore, to avoid risky situations caused by a single low-quality decision, it is encouraged to enhance the decision-making robustness by directly inducing the VLM-based decision-maker to recognize the presence of randomness during inference, generating multiple candidate decisions awaiting arbitration until more information from the planning module is obtained.

Under the guidance of \Cref{tab:decision_guideline}, the reasoning process of our VLM-based decision-maker is structured to emulate the human cognitive process of \textit{condition analysis}, \textit{hypothesis formulation}, and \textit{self-evaluation}~\cite{newell1972human, koriat1993we}. During the process, scene understanding is leveraged as a foundational step similar to the practice in related knowledge-driven approaches~\cite{sima2024drivelm, zheng2024planagent}, where the VLM is instructed to capture relevant information in the given scenario to support the subsequent decision reasoning. We decompose the action selection and confidence assessment into separate reasoning stages, with the motivation of allowing the VLM to better evaluate confidences after every possible solution with the rationale is clearly presented in the responding context. The structured reasoning process and outputs are formally expressed as follows:
\begin{subequations}
    \begin{align}
    \bm{R}_{1} &= \mathcal{G}_1(\bm{I}_{t}, \bm{D}_t), \label{subeq: scene understand}\\
    \bm{R}_{2} &= \mathcal{G}_2(\bm{I}_{t}, \bm{D}_t, \bm{R}_{1}), \label{subeq: candidate select}\\
    \bm{R}_{3} &= \mathcal{G}_3(\bm{I}_{t}, \bm{D}_t, \bm{R}_{1}, \bm{R}_{2}), \label{subeq: confidence assess}\\
    \bm{R}_{2} \Rightarrow \bm{A}_t, \ &\bm{A}_t = \left[\bm{a}^1_t,\bm{a}^2_t,...,\bm{a}^K_t\right], \label{subeq: action represent} \\
    \bm{R}_{3} \Rightarrow \bm{c}_t, \ &\bm{c}_t = \left[c^1_t,c^2_t,...,c^K_t\right], \label{subeq: confidence represent} 
\end{align}
\end{subequations}
where \(\mathcal{G}_1\), \(\mathcal{G}_2\), and \(\mathcal{G}_3\) denote the VLM inference at the stage of scene understanding, action selection, and confidence assessment, respectively, with \(\bm{R}_1\), \(\bm{R}_2\), and \(\bm{R}_3\) corresponding to the responding contents. Derived from \(\bm{R}_2\), the elements in $\bm{A}_t$ are the selected Top-K actions in action space $\mathcal{A}$, and each scalar in $\bm{c}_t$ extracted from \(\bm{R}_3\) represents a confidence level lying in $\left[0, 1\right]$.

\begin{table*}[htbp]
\centering
\caption{Decision-Making Guideline. S.U.: Scene Understanding. A.S.: Action Selection. C.A.: Confidence Assessment.}
\label{tab:decision_guideline}
\renewcommand{\arraystretch}{1.0} 
\begin{tabular}{>{}p{3cm}p{\dimexpr\textwidth-3cm-4\tabcolsep}}
\toprule
\multicolumn{1}{l}{\textbf{Component}} & \multicolumn{1}{l}{\textbf{Description in Human Message}} \\
\midrule
Goal Explanation &
The driving goal is to be `as efficient and comfortable as possible while ensuing safety'. To achieve this, your reasoning goes through the following steps.\\
\midrule
S.U. Instruction & 
Step 1 - scene understanding: analyze the driving context, E.g. road condition, intention of surrounding road users, navigation for the ego vehicle (if applicable). \\
\midrule
A.S. Instruction & 
Step 2 - action selection: based on the Step-1 analysis and common-sense knowledge, infer your Top-\{K\} answers in the \{action\_set\}. \\
\midrule
C.A. Instruction & 
Step 3 - confidence assessment: explicitly analyze the efficiency, comfort, and safety of each candidate action, then rate your confidence that each choice will meet the goal. \\
\bottomrule
\end{tabular}
\vspace{-0.5em}
\end{table*}

\subsection{Guided Trajectory Generation and Hierarchical Refinement}
\subsubsection{Objective Creator}\label{sec:objective creator}
The objective creator maps each candidate decision made by the VLM to the corresponding objective function to guide the downstream motion planning. The mapping is achieved by multiplying a decision-following objective with a basic objective that represents the general goodness of a trajectory, expressed as:
\begin{equation}\label{eq:objective}
    J^k = (J^k_f)^{\omega_f} \cdot (J_g)^{\omega_g}, \forall k \in \left\{1,2,...,K\right\},
\end{equation}
where $J^k_{f}$ represents the decision-following objective of the $k$-th candidate decision $a^k_t$, and $J_g$ represents the general objective. Both $J^k_{f}$ and $J_g$ lie in $\left[0, 1\right]$. $\omega_f,\omega_g \in \mathbb{R}^+$ 
are tunable parameters for controlling the relative importance of the two sub-objectives. 

We formulate the decision-following objective as:
\begin{small}
    \begin{equation}\label{eq:dec follow}
    \begin{aligned}
        J^k_f &= \underbrace{\max \left(1 - \frac{1}{T} \cdot \sum^{t+T}_{\tau=t} \left\| \bm{x}_{\tau} - \left(\bm{x}^r\right)^k_{\tau} \right\|_2 \cdot \frac{1}{d_{max}}, 0\right)}_{\text{Lane Following}}\\
        &\cdot \underbrace{\max \left(1 - \frac{1}{T} \cdot \sum^{t+T}_{\tau=t} \mathds{1}\left(v_{\tau} \notin \mathcal{V}^k \right) \inf_{v\in\mathcal{V}^k}
        \left\|v_{\tau} - v \right\|_1 \cdot \Delta t, 0\right)}_{\text{Speed Following}},
    \end{aligned}
\end{equation}
\end{small}
where $T$ and $\Delta t$ represent the planning horizon and time resolution, respectively. $\bm{x}_{\tau}$ and $v_{\tau}$ denote the ego coordinates and ego speed at time-step $\tau$. $\left(\bm{x}^r\right)^k_{\tau}$ is the coordinates of the reference waypoint on the target lane centerline, which can be queried with the high-definition map or perception outputs. $d_{max}$ is a constant indicating the maximum acceptable lateral deviation. \(\mathds{1}\) represents the indicator function. $\mathcal{V}^k$ stands for the reference speed interval, calculated based on the current speed and longitudinal action:
\begin{equation}\label{eq:speed interval}
 \mathcal{V}^k(\bm{a}^k_t, v_t) := 
\begin{cases}
\phantom{\{} [\max(\lambda \cdot v_t, \ \bar{v}), \ +\infty), & \text{if} \ a_{acc} \in \bm{a}^k_t, \\
\phantom{\{} [\gamma \cdot v_t, \ \max(\lambda \cdot v_t, \ \bar{v})), &  \text{if} \ a_{cruise} \in \bm{a}^k_t, \\
\phantom{\{} [0, \ \gamma \cdot v_t), & \text{if} \ a_{dec} \in \bm{a}^k_t,
\end{cases}
\end{equation}
where \(\lambda > 1\) and \(0<\gamma < 1\), serving as scaling parameters for the calculation of higher and lower reference speeds. The constant \(\bar{v}\) is defined as a speed threshold to ensure the lower bound of the higher reference speed, which is introduced to prevent the inability of accelerating when \(v_t\) is in the vicinity of 0. Overall, the definition of \(\mathcal{V}^{k}\) enables speed guidance conforming to the semantics of longitudinal decisions by penalizing planned speeds based on their deviations from the prescribed interval. 

For the general objective $J_g$, we adopt the formulation of Predictive Driver Model (PDM) scorer~\cite{dauner2023parting}:
\begin{equation}\label{eq:general obj}
    J_g = \prod_{c \in \mathcal{C}} m_{c} \cdot \frac{\sum_{s \in \mathcal{S}} \omega_{s} \cdot m_{s}}{\sum_{s \in \mathcal{S}} \omega_s},
\end{equation}
where \(\mathcal{C}\) denotes a set of constraints and \(\mathcal{S}\) represents a set of soft metrics. \(m_c, m_s \in \left[0, 1\right]\) are the scores of each sub-term. \(w_s \in \mathbb{R}^+\) is the weight of each soft metric. Specifically, our implementation of \(J_g\) references the modified PDM scorer proposed by~\cite{yang2024diffusion}, where \(\mathcal{C}\) incorporates collision avoidance, driving direction compliance, drivable area compliance, and speed limit compliance, and \(\mathcal{S}\) synthesizes progress, time-to-collision within bound, comfort, centerline following, and gap to the leading vehicle.

\begin{remark}\label{remark: hard constraints}
    \(\forall c \in \mathcal{C}\), multiplier score \(m_c\) is defined as 0 when the trajectory severely violates the constraint (\eg, at-fault collision, off-road driving)~\cite{yang2024diffusion}. The multiplicative formulation ensures that trajectories with critical violations receive zero scores, effectively filtering infeasible options. By integrating \(J_g\) into Eq.~\eqref{eq:objective}, the framework imposes explicit constraints regarding safety and traffic rule compliance when translating high-level decisions to executable trajectories.
\end{remark}

\subsubsection{Gradient-Free Diffusion-Based Motion Planning}\label{sec:gradient free plan}
With $K$ objectives created, we next perform decision-guided trajectory generation and first-stage refinement:
\begin{equation}\label{eq:fist refine}
        (\bm{X}^k)^{*} = \argmax_{\bm{X}^k_n \in \mathcal{X}^k} J^k(\bm{X}^k_n), \forall k \in \{1,2, ...,K\},
\end{equation}
where $\mathcal{X}^k$ denotes the set of trajectory proposals generated under decision $k$, with $\left|\mathcal{X}^k\right|= N$. Each trajectory proposal \(\bm{X}^k_n \in \mathbb{R}^{3*T}\) comprises a time sequence of SE(2) poses \(\left(\bm{x}_{\tau}, \psi_{\tau}\right)\), where \(\psi_{\tau}\) denotes the ego heading at time-step \(\tau\). The highest-scored proposal $(\bm{X}^k)^{*}$ is selected as the output. 
To effectively explore the two-dimensional trajectory space and generate sufficient proposals for each decision, we leverage the strong generative capacity of diffusion models, which excel at producing high-quality trajectory proposals while maintaining flexibility in motion planning.
Specifically, we adopt \textit{Diffusion-ES}~\cite{yang2024diffusion} to obtain trajectory proposals under each decision, which adapts to an arbitrary form of objectives with an evolutionary-strategy-enabled gradient-free optimization. An iterative process of denoising, scoring, sampling, and renoising is performed to mutate trajectories towards desired proposals:
\begin{subequations}\label{eq:diffusiones}
    \begin{align}
        \mathcal{X} &= denoise(\Tilde{\mathcal{X}};\theta) \label{subeq: init denoise} ,\\
        q(\bm{X}_n) & = \frac{\exp(\mu \cdot J^k(\bm{X}_n))}{\sum_{\bm{X}_i \in \mathcal{X}} \exp(\mu \cdot J^k(\bm{X}_i))}, \forall \bm{X}_n \in \mathcal{X}, \label{subeq: sample prob} \\
        \mathcal{Y} &= \{ \bm{Y}_n \mid \bm{Y}_n \sim q(\bm{X}_n), \bm{X}_n \in \mathcal{X} \}, \label{subeq: elite sample} \\
        \Tilde{\mathcal{Y}} &= renoise(\mathcal{Y}; \epsilon, \beta_j), \epsilon \sim \mathcal{N}(\bm{0}, \bm{1}), \beta_j \in (0,1),\\
        \mathcal{X}^k &= denoise(\Tilde{\mathcal{Y}};\theta), \label{subeq: truncated denoise}
    \end{align}
\end{subequations}
where Eq.~\eqref{subeq: init denoise} first denoises randomized input $\Tilde{\mathcal{X}}$ with trained parameters $\theta$ and obtains clean samples $\mathcal{X}$. A subsequent sampling procedure adopts Eq.~\eqref{subeq: sample prob} to calculate the probability of drawing each clean sample, $q(\bm{X}_n)$, according to their scores measured by the objective function, where $\mu$ serves as a temperature parameter. Since higher-scored trajectories are associated with higher sampling probability, a set of \textit{elite} trajectories is collected as denoted by $\mathcal{Y}$ in Eq.~\eqref{subeq: elite sample}, with $\left|\mathcal{Y}\right|=N$. These trajectories are renoised into $\Tilde{\mathcal{Y}}$ with a variance schedule $\beta_j \in (0,1)$, where $j$ is a step of adding Gaussian noise. Finally, a denoising process recovers $\Tilde{\mathcal{Y}}$ into $\mathcal{X}^k$, a set of decision $k$-guided proposals. Eq.~\eqref{subeq: sample prob} - \eqref{subeq: truncated denoise} run iteratively to offer high-quality proposals without gradient information.

\begin{algorithm}
\caption{Decision-Guided Trajectory Generation and Hierarchical Refinement}
\label{alg:guidance_refinement}
\begin{algorithmic}[1]
\STATE \textbf{Input:} Top-K Decisions $\bm{A}_t = \left[\bm{a}_t^1,\bm{a}^2_t, \dots, \bm{a}_t^K\right]$ and Confidences $\bm{c}_t = \left[c_t^1,c_t^2, \dots, c_t^K\right]$ from VLM; Current Ego State $\bm{s}_t = \left[\bm{x}_t, v_t, \psi_t\right]$; Environment Context $\mathcal{E}_t$.

\STATE \textbf{Initialize:} $S^* \leftarrow -\infty$, $\bm{X}^* \leftarrow \emptyset$.

\FOR{$k = 1$ \TO $K$}
    \STATE \textit{// 1. Objective Creation (Sec.~\ref{sec:objective creator})}
    \STATE Determine reference speed interval $\mathcal{V}^k(\bm{a}_t^k, v_t)$ $\triangleright$ \textbf{Eq.~\eqref{eq:speed interval}}.
    \STATE Define decision-following objective $J_f^k(\cdot)$ using $\mathcal{V}^k$ and path reference from \(\mathcal{E}_t\) $\triangleright$ \textbf{Eq.~\eqref{eq:dec follow}}.
    \STATE Define guidance objective: 
    \STATE \quad $J^k \leftarrow (J_f^k)^{\omega_f} \cdot (J_g)^{\omega_g}$ $\triangleright$ \textbf{Eq.~\eqref{eq:objective}}.

    \STATE \textit{// 2. Guided Generation \& First-Stage Refinement (Sec.~\ref{sec:gradient free plan})}
    \STATE Sample initial noise trajectories: $\tilde{\mathcal{X}} \sim \mathcal{N}(0, I)$.
    \STATE Generate candidate set $\mathcal{X}^k$ via gradient-free optimization over $J^k$: 
    \STATE \quad $\mathcal{X}^k \leftarrow \text{Diffusion-ES}(\tilde{\mathcal{X}}, J^k; \theta)$. \hfill $\triangleright$ Iterative process in \textbf{Eq.~\eqref{eq:diffusiones}}
    \STATE Choose the best proposal for decision $k$:
    \STATE \quad $(\bm{X}^k)^* \leftarrow \operatorname*{arg\,max}_{\bm{X}_n^k \in \mathcal{X}^k} J^k(\bm{X}^k_n)$. \hfill $\triangleright$ \textbf{Eq.~\eqref{eq:fist refine}}
    
    \STATE \textit{// 3. Second-Stage Refinement (Sec.~\ref{sec:confidence_aware_selection})}
    \STATE Evaluate with preference awareness:
    \STATE \quad $\tilde{J}^k((\bm{X}^k)^*) \leftarrow (J_f^k((\bm{X}^k)^*))^{\tilde{\omega}_f} \cdot (J_g((\bm{X}^k)^*))^{\tilde{\omega}_g}$.
    \STATE \quad $S((\bm{X}^k)^*) \leftarrow (c_t^k)^{\omega_c} \cdot \tilde{J}^k((\bm{X}^k)^*)$. \hfill $\triangleright$ \textbf{Eq.~\eqref{eq: conf-aware select}}

    \IF{$S((\bm{X}^k)^*) > S^*$}
        \STATE $S^* \leftarrow S((\bm{X}^k)^*)$.
        \STATE $\bm{X}^* \leftarrow (\bm{X}^k)^*$.
    \ENDIF
\ENDFOR
\STATE \textbf{return} final trajectory plan $\bm{X}^*$.
\end{algorithmic}
\end{algorithm}

\subsubsection{Preference-Informed Trajectory Selection}\label{sec:confidence_aware_selection}
The second-stage refinement involves selecting the best trajectory for tracking among $K$ multimodal candidates generated by the motion planner. We utilize the following expression capturing both the preference of decision-making and the quality of motion planning to score the candidates:
\begin{equation}\label{eq: conf-aware select}
    \begin{aligned}
        S\left((\bm{X}^k)^*\right) &= \left(c^k_t\right)^{\omega_{c}} \cdot \tilde{J}^k\left((\bm{X}^k)^*\right), \\
        \tilde{J}^k &= (J^k_f)^{\tilde{\omega}_f} \cdot (J_g)^{\tilde{\omega}_g}, \forall k \in \left\{1,2,...,K\right\},
    \end{aligned}
\end{equation}
where $\omega_c \in \mathbb{R}^+$ controls the relative importance of the VLM's decision-making confidence in the comprehensive scorer. $\tilde{J}^k$ is calculated identically to $J^k$ with only differences in parameter settings, which allows for different trajectory selection criteria. We adopt a greater $\omega_f$ to encourage decision following in the first stage and choose a smaller $\tilde{\omega}_f$ for balancing with general quality in the second stage.

We use $c_t^k$ as a verbalized confidence score that operationally expresses the VLM's relative preference strength among the elicited tactical alternatives. Accordingly, Eq.~\eqref{eq: conf-aware select} is an empirical multiplicative fusion rule: $c_t^k$ supplies a heuristic arbitration signal, whereas $\tilde{J}^k$ measures the executable trajectory using explicit planning objectives. The exponent $\omega_c$ controls the influence of the preference signal. 
The incremental utility of this formulation is evaluated empirically in \Cref{sec:confidence analysis}.

\subsubsection{Overall Algorithm Implementation}
Algorithm~\ref{alg:guidance_refinement} integrates the three core components described in Sections~\ref{sec:objective creator} to \ref{sec:confidence_aware_selection} into a unified framework for decision-guided trajectory generation and hierarchical refinement. The algorithm assumes access to a pre-trained diffusion model \(\theta\) and all the hyper-parameters specified. It receives the Top-K decisions \(\bm{A}_t\) and confidences \(\bm{c}_t\) from the VLM module, the current ego state \(\bm{s}_t\), and environment context $\mathcal{E}_t$ (including surrounding agents' states and lane geometry) from the perception system, generating the final trajectory plan \(\bm{X}^*\) ready for execution by the controller.

\begin{table}[t]
\caption{Parameter settings in our implementation.}
\centering
\renewcommand{\arraystretch}{1.5}
\resizebox{0.8\linewidth}{!}{
\begin{tabular}{l|l|l}
\hline
\textbf{Notation} & \textbf{Parameter} & \textbf{Value}\\ \hline
\(\bar{d}\) & Distance for scenario division & 20.0\,m\\
\(C_d\) & Cycle of decision-making  & 2.0\,s \\
\(C_p\)  & Cycle of motion planning & 0.5\,s \\ 
\(\Delta t\)  & Time-step resolution & 0.1\,s \\
\(T\)  & Horizon of motion planning & 40  \\
\(K\) & Num. candidate decisions & 3\\
\(-\) & Temperature of VLM & 0.0\\
\(\lambda\) & Coefficient of higher ref. speed & 1.25\\
\(\gamma\) & Coefficient of lower ref. speed & 0.75\\
\(\bar{v}\) & Minimum higher ref. speed & 2.0\,\text{m/s}\\
\(d_{max}\) & Maximum allowed lateral deviation & 5.0\,m\\
\(\omega_c\) & Power parameter of \(c^k_t\) & 1.0\\
\(\omega_f\) & Power parameter  of \(J^k_f\) in Eq.~\eqref{eq:objective} & 5.0\\
\(\tilde{\omega}_f\) & Power parameter of \(J^k_f\) in Eq.~\eqref{eq: conf-aware select} & 0.3\\
\(\omega_g\) & Power parameter  of \(J^k_g\) in Eq.~\eqref{eq:objective} & 1.0\\
\(\tilde{\omega}_g\) & Power parameter of \(J^k_g\) in Eq.~\eqref{eq: conf-aware select} & 1.0\\
\end{tabular}
}
\vspace{-0.5em}
\label{tab:para setting}
\end{table}

\section{Experiment}

\begin{table*}
    \centering
    \caption{Quantitative closed-loop performance of different approaches on nuPlan and InterPlan. All metrics are \underline{higher the better}; the best and second-best planner scores are highlighted in \colorbox{lightblue}{light blue} and \second{pale yellow}, respectively. Log-Replay is an expert reference and is excluded from ranking. \(*\): Metrics quoted from original publications. \dag: Methods with rule-based modules for emergency braking. \ddag: Our approach with long-term memory module.}
    \vspace{-0.5em}
    \resizebox{0.9\textwidth}{!}{
    \begin{tabular}{@{}llcccccccc|cc@{}}  
        \toprule
        \multirow{2}{*}{\textbf{Type}} & \multirow{2}{*}{\textbf{Planner}} 
        & \multicolumn{4}{c}{\textbf{nuPlan Test14-Hard}} 
        & \multicolumn{4}{c}{\textbf{nuPlan Test14-Random}} 
        & \multicolumn{2}{c}{\textbf{InterPlan}} \\  
        \cmidrule(lr){3-6} \cmidrule(lr){7-10} \cmidrule(lr){11-12}  
        & & \textbf{NR-CLS}  & \textbf{NR-SR(\%)}  & \textbf{R-CLS}  & \textbf{R-SR(\%)}  & \textbf{NR-CLS}  & \textbf{NR-SR(\%)}  & \textbf{R-CLS}  & \textbf{R-SR(\%)} & \textbf{R-CLS}  & \textbf{R-SR(\%)} \\
        \midrule
        \multirow{2}{*}{Rule-Based} & IDM~\cite{helbing1998generalized} & 56.16 & 65.81 & 62.26 & 71.69 & 70.39 & 77.01 & 72.42 & 78.54 & 47.08 & 58.51 \\
        & PDM-Closed\(^{\dag}\)~\cite{dauner2023parting} & 65.08 & 78.68 & 75.19 & 85.66 & 90.05 & 94.64 & \makecell{\colorbox{lightblue}{91.64}} & \makecell{\second{96.55}} & 69.67 & 83.58 \\
        \midrule
        \multirow{7}{*}{Learning-Based} 
                                                 & UrbanDriverOL~\cite{scheel2022urban} & 51.67 & 59.20 & 49.06 & 55.51 & 63.57 & 70.88 & 60.92 & 66.28 & 15.17 & 21.49 \\
                                                 & RasterModel~\cite{caesar2021nuplan} & 50.65 & 58.09 & 52.44 & 59.56 & 67.70 & 73.56 & 68.64 & 74.33 & 35.54 & 48.96 \\
                                                 & PlanTF~\cite{cheng2024rethinking} & 72.56 & 79.04 & 60.34 & 67.28 & 85.60 & 90.80 & 78.86 & 85.44 & 47.69 & 61.79 \\
                                                 & Diffusion-Planner~\cite{zhengdiffusion} & 75.29 & 82.72 & 68.83 & 78.31 & 88.98 & 93.87 & 83.21 & 90.42 & 53.04 & 69.25 \\
                                                 & UniPlanner\(^*\)~\cite{yang2025uniplanner} & 71.38 & - & 70.99 & - & 87.25 & - & 85.25 & - & - & - \\
                                                 & CausalPlanner\(^*\)~\cite{xie2026causalplanner} & 76.30 & - & 67.39 & - & \makecell{\second{90.99}} & - & 84.80 & -  & 60.76 & -\\
                                                 & Flow-Planner~\cite{tan2026flow} & 73.68 & 81.25 & 67.49 & 75.37 & 85.79 & 90.04 & 81.54 & 87.36 & 54.57 & 70.75 \\
        \midrule
        \multirow{4}{*}{Hybrid} & PDM-Hybrid\(^{\dag}\)~\cite{dauner2023parting} & 65.99 & 79.41 & 76.07 & 86.40 & 90.10 & 95.02 & \makecell{\second{91.29}} & 96.17 & 69.44 & 83.58 \\
                                         & GameFormer~\cite{huang2023gameformer} & 62.94 & 74.26 & 61.53 & 72.79 & 76.89 & 85.29 & 77.80 & 86.59 & 46.77 & 62.09 \\
                                         & Diffusion-ES~\cite{yang2024diffusion} & \makecell{\second{77.54}} & 84.93 & 77.75 & 87.13 & 87.70 & 94.25 & 87.18 & 93.87 & 66.19 & 84.18 \\
                                         & PLUTO\(^{\dag}\)~\cite{cheng2024pluto} & \makecell{\colorbox{lightblue}{79.19}} & 88.24 & 75.75 & 84.19 & \makecell{\colorbox{lightblue}{91.93}} & \makecell{\second{95.79}} & 89.94 & 96.17 & 63.94 & 79.10 \\
        \midrule
        \multirow{6}{*}{Knowledge-Driven} & PlanAgent\(^*\)~\cite{zheng2024planagent} & 72.51 & - & 76.82 & - & - & - & - & - & - & - \\ 
                                    & DualAD (GLM-4-Flash)~\cite{wang2025dualad} & 63.38 & 73.16 & 66.40 & 76.10 & 78.69 & 86.21 & 78.28 & 84.67 & 51.62 & 61.79 \\
                                    & DualAD (GPT-4o)~\cite{wang2025dualad} & 60.87 & 72.43 & 64.57 & 76.10 & 78.30 & 87.36 & 80.67 & 89.66 & 52.81 & 63.88 \\
                                    & PRIMA (Ours) & 77.39 & \makecell{\second{88.97}} & 78.13 & \makecell{\second{89.71}} & 86.70 & 95.02 & 87.11 & 95.79 & 69.23 & 86.57 \\
                                    & PRIMA\(^{\ddag}\) (Ours) & 75.03 & \makecell{\colorbox{lightblue}{90.44}} & \makecell{\second{78.46}} & \makecell{\colorbox{lightblue}{90.07}} & 87.71 & \makecell{\colorbox{lightblue}{98.47}} & 88.08 & \makecell{\colorbox{lightblue}{96.93}} & \makecell{\second{70.11}} & \makecell{\second{87.46}} \\
                                    & PRIMA-Lite (Ours) & 67.89 & 83.82 & 77.26 & \makecell{\colorbox{lightblue}{90.07}} & 85.12 & 95.02 & 89.13 & \makecell{\colorbox{lightblue}{96.93}} & 67.69 & 84.78 \\
        \midrule
        Multi-Planner & SAH-Drive~\cite{fan2025sah} & 70.58 & 84.19 & \makecell{\colorbox{lightblue}{78.51}} & \makecell{\second{89.71}} & 87.22 & 95.02 & 89.79 & 95.79 & \makecell{\colorbox{lightblue}{78.30}} & \makecell{\colorbox{lightblue}{94.63}} \\
        \midrule
        \textcolor{mediumgray}{Expert} & \textcolor{mediumgray}{Log-Replay} & \textcolor{mediumgray}{85.96} & \textcolor{mediumgray}{91.18} & \textcolor{mediumgray}{68.80} & \textcolor{mediumgray}{73.53} & \textcolor{mediumgray}{94.03} & \textcolor{mediumgray}{96.93} & \textcolor{mediumgray}{75.86} & \textcolor{mediumgray}{78.16} & -  & - \\
        \bottomrule
    
\end{tabular}
    \label{tab:planner compare}
    }
\vspace{-1em}
\end{table*}

\begin{table*}[t]
\centering
\caption{InterPlan breakdown on the same 335 scenarios. All entries are official mean scores in percent; higher is better. Light blue boxes indicate the best value in each column. The learning and PDM rows fix the corresponding SAH-Drive outputs, whereas switch denotes its original adaptive configuration. The first four submetrics are multiplicative under the default scenario protocol; InterPlan adds scenario-specific passing multipliers and a weighted goal-lane-change score.}
\vspace{-0.5em}
\label{tab:interplan_breakdown}
\resizebox{0.9\textwidth}{!}{{
\begin{tabular}{@{}lcc|cccc|ccc|c@{}}
\toprule
& \multicolumn{2}{c|}{Overall} & \multicolumn{4}{c|}{Default multiplicative metrics} & \multicolumn{3}{c|}{InterPlan passing metrics} & InterPlan weighted \\
Planner & R-CLS & R-SR(\%) & No collision & Drivable area & Making progress & Driving direction & Construction & Stopped vehicle & Jaywalker & Goal-lane changes \\
& (335) & (335) & (335) & (335) & (335) & (295) & (10) & (30) & (10) & (335) \\
\midrule
PRIMA$^{\ddag}$ & 70.11 & 87.46 & \colorbox{lightblue}{98.81} & \colorbox{lightblue}{99.40} & 92.54 & 99.15 & 50.00 & 33.33 & \colorbox{lightblue}{100.00} & \colorbox{lightblue}{76.24} \\
SAH-Drive (learning) & 71.15 & 90.15 & 98.51 & 92.24 & \colorbox{lightblue}{98.21} & 98.47 & \colorbox{lightblue}{100.00} & \colorbox{lightblue}{86.67} & \colorbox{lightblue}{100.00} & 73.28 \\
SAH-Drive (PDM) & 70.95 & 85.07 & 95.82 & 98.21 & 93.73 & \colorbox{lightblue}{100.00} & 20.00 & 36.67 & \colorbox{lightblue}{100.00} & 71.95 \\
SAH-Drive (switch) & \colorbox{lightblue}{78.30} & \colorbox{lightblue}{94.63} & 98.21 & 98.81 & 97.91 & 99.66 & 90.00 & 83.33 & \colorbox{lightblue}{100.00} & 71.73 \\
\bottomrule
\end{tabular}}}
\parbox{0.99\textwidth}{\scriptsize{Numbers in parentheses give the number of applicable scenarios. Driving-direction compliance is replaced by obstacle-passing metrics in construction and vehicle-obstacle scenarios; jaywalking retains it. The four jaywalker scores are tied. Submetric means are descriptive and do not add up to R-CLS.}}
\vspace{-0.5em}
\end{table*}

\subsection{Simulation Setup}
We evaluate the proposed PRIMA on nuPlan~\cite{caesar2021nuplan} and InterPlan~\cite{hallgarten2024can}.

\textbf{nuPlan.} NuPlan serves as a well-established simulation platform built upon various real-world traffic scenarios for systematic driving performance testing. \textbf{Test14-Hard} and \textbf{Test14-Random} benchmarks~\cite{cheng2024rethinking} are utilized for our evaluation. Both of them are developed to encompass 14 scenario types in the nuPlan Planning Challenge, and the former is especially introduced to represent \textit{long-tail} scenarios. The experiment is conducted in both \textbf{Non-Reactive} and \textbf{Reactive} modes. Under the non-reactive setting, the motion of all traffic participants is simulated with log-replay. In the reactive mode, Intelligent-Driver-Model (IDM)~\cite{helbing1998generalized} is deployed for the simulation of vehicle motion. We adopt official metrics including Non-Reactive Closed-Loop Score (\textbf{NR-CLS}) and Reactive Closed-Loop Score (\textbf{R-CLS}) for the comprehensive performance assessment integrating dimensions such as efficiency, comfort, human-likeness, \etc Additionally, Non-Reactive Success Rate (\textbf{NR-SR}) and Reactive Success Rate (\textbf{R-SR}) are leveraged to indicate the ability of a method to avoid severe errors, calculated as the percentage that the method does not get a closed-loop score of \textit{zero} due to collision, out of drivable area, incorrect driving direction, or low progress~\cite{caesar2021nuplan}.

\textbf{InterPlan.}
InterPlan is constructed to feature highly interactive and long-tail edge cases, evaluating cross-domain generalization and robustness against distributional shifts and environmental disturbances. 
We leverage the official \textbf{benchmark scenarios} comprising 335 driving tasks, which are derived from real-world data but enriched with rare situations (\eg, construction zones) and diverse agent policies.
Our evaluation follows the official extended R-CLS~\cite{hallgarten2024can}. The metric accommodates long-tail scenarios by incorporating lane-change completion rates, applying multiplicative progress penalties to deter stalling, and exempting wrong-way violations during overtakes or accident avoidance.
The definition of R-SR remains as the percentage of scenarios yielding a non-zero score to quantify the avoidance of critical failures.

\textbf{Implementation Details.}  For all simulations, a Linear-Quadratic-Regulator tracker is employed to map the planned trajectory to control inputs, which are sent to a kinematic bicycle model to propagate the dynamics. 
The simulations are conducted in synchronous mode with simulator pausing to bypass stale-decision handling, isolating the test of the framework's cognitive driving capability~\cite{wendilu, zheng2024planagent}.
In framework implementation, we prioritize using GPT-4o as the VLM to build our decision-making module. The Diffusion-ES is trained and deployed following the efficient version specified by~\cite{yang2024diffusion}, where the number of denoising steps is given as 10 and the number of renoising-denoising iterations is set to 2, with the only difference that we choose a proposal number of 128 instead of 32. Consequently, our framework encompasses 6.72M trainable parameters of the diffusion-based planner, while instantiating the VLM as a frozen API-based reasoning engine.
Other parameter settings are reported in \Cref{tab:para setting}.

\subsection{Quantitative Studies}
\label{sec:quantitative}
We compare PRIMA with rule-based, learning-based, hybrid, knowledge-driven, and multi-planner approaches under the same nuPlan and InterPlan protocols. 
For fair comparison, we benchmark the open-source implementation of different methods on a four-card RTX-4090 computing device in our evaluation. For methods remaining closed-source, the results are quoted from the original publications. 

To fully leverage the knowledge-driven nature of our pipeline and enhance closed-loop adaptability, we introduce a retrieval-augmented variant, PRIMA$^{\ddag}$, which further incorporates a long-term memory module~\cite{wendilu}. Specifically, we deploy PRIMA on the nuPlan validation split and collect 626 decision-making demonstrations from the highest-scored scenarios among the 14 challenging scenario types. The driving contexts of these samples are multimodally encoded via Qwen3-VL-Embedding integrating BEV images and textual descriptions. During inference,  PRIMA$^{\ddag}$ retrieves the top-2 similar demonstrations via cosine similarity and injects them as few-shot context, allowing prior successful experiences to guide the current decision-making.

To examine a lower-latency implementation under the same closed-loop protocols, we additionally evaluate PRIMA-Lite. Its Qwen3-VL-2B student model is distilled from the same 626 teacher-labeled nuPlan-validation examples that form PRIMA$^{\ddag}$'s memory library; these examples are disjoint from the evaluated nuPlan and InterPlan scenarios. The student model directly predicts the top-three tactical labels without generating a chain of thought. For planning, we construct an offline, speed-conditioned library of 8192 Diffusion-ES trajectories (eight speed groups of 1024). At each step, the three groups nearest the current ego speed are searched, and 128 candidates per action are shortlisted using route deviation and kinematic penalties before kinematic rollout and trajectory scoring.

Quantitative results are summarized in \Cref{tab:planner compare}, from which we present the following findings:

\noindent \textbf{1. PRIMA$^{\ddag}$ leads Test14-Hard in NR-SR and R-SR and ranks second in R-CLS, only 0.05 points behind SAH-Drive.} This combination of high success rates and strong closed-loop scores reflects robust performance on challenging long-tail cases.

\noindent \textbf{2. With long-term memory, PRIMA$^{\ddag}$ reaches 98.47\% NR-SR and 96.93\% R-SR on Test14-Random, and a competitive 70.11 R-CLS/87.46\% R-SR on InterPlan.} The InterPlan R-CLS and R-SR rank second in \Cref{tab:planner compare}, following SAH-Drive. The metric breakdown in \Cref{tab:interplan_breakdown} shows complementary safety and goal-lane strengths.

\noindent \textbf{3. PRIMA improves on Diffusion-ES across all three evaluated benchmarks.} Compared to Diffusion-ES, a SOTA generation-then-scoring planner, our base model improves NR-SR and R-SR by 4.04\% and 2.58\% on Test14-Hard, and by 0.77\% and 1.92\% on Test14-Random, achieving dual-metric improvements on InterPlan (+3.04 R-CLS, +2.39\% R-SR). These consistent SR gains motivate the controlled analysis in \Cref{sec:confidence analysis}, which isolates the contribution of preference-informed arbitration from other pipeline components.
Compared to the recent two-stage framework UniPlanner, PRIMA$^{\ddag}$ shows better performance across all test settings. 

\noindent \textbf{4. PRIMA outperforms all knowledge-driven baselines without requiring online reflection and re-decision iterations.} Compared to PlanAgent, our Top-K confidence elicitation enables multiple candidate decisions in a single dialog round, accounting for downstream planning inconsistencies. DualAD directly outputs reference speeds with lower computational cost but obtains lower CLS on the evaluated benchmarks. PRIMA-Lite demonstrates that direct inference without CoT can also achieve favorable performance when it retains Top-3 tactical alternatives and downstream trajectory evaluation.

\noindent \textbf{5. PRIMA-Lite offers strong reactive performance.} It reaches 77.26 R-CLS/90.07\% R-SR on Test14-Hard and 89.13 R-CLS/96.93\% R-SR on Test14-Random, tying the best R-SR on both nuPlan tests. On Test14-Random NR, it remains close to PRIMA (85.12 versus 86.70 NR-CLS; both 95.02\% NR-SR). However, Test14-Hard NR and InterPlan remain weaker than the original PRIMA variants, indicating that the current lower-latency configuration does not yet maintain the same performance across all settings.

\begin{figure*}
  \centering
    \includegraphics[width=1\linewidth]{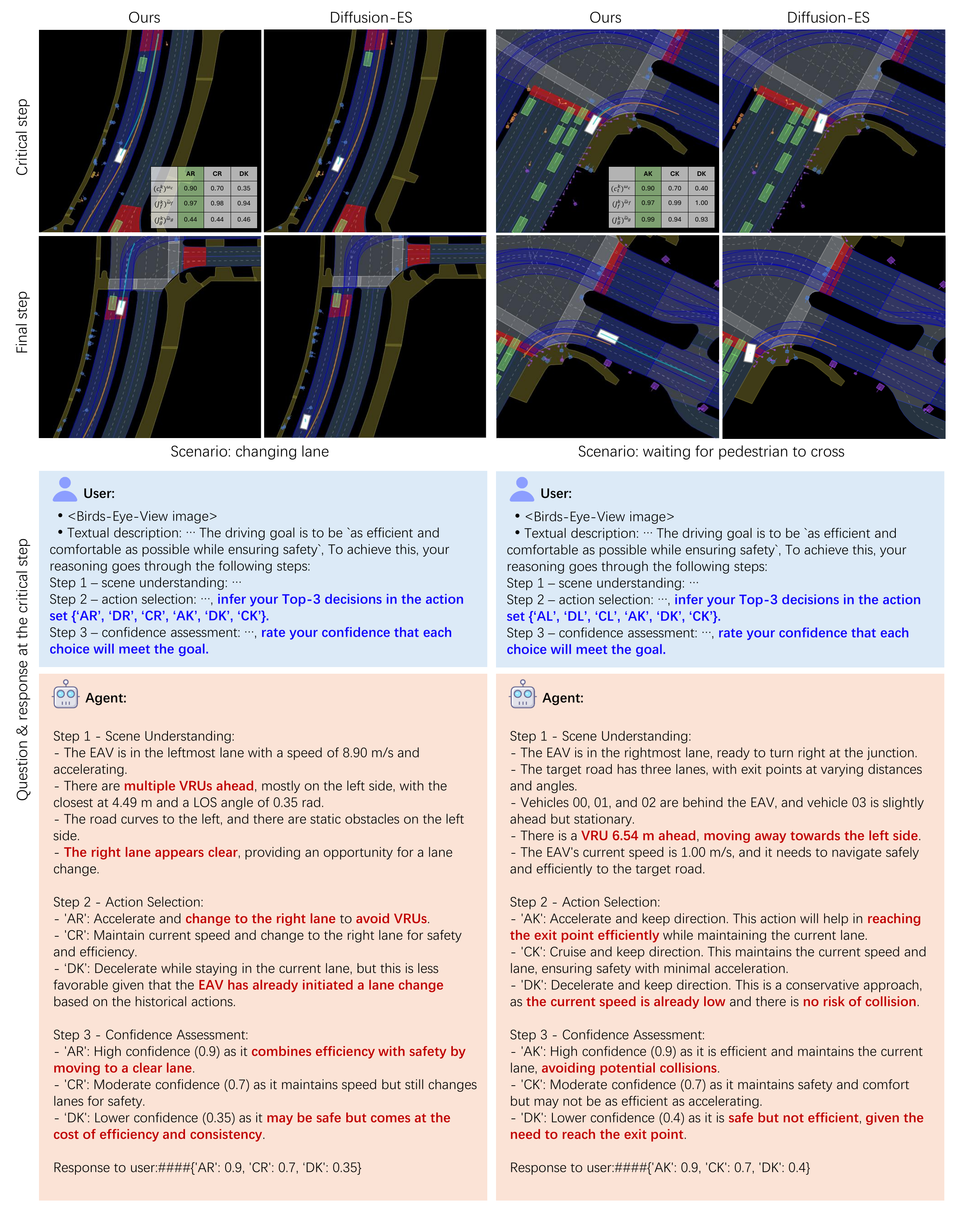}
  \caption{A comparison between our approach and Diffusion-ES in two representative driving scenarios. At the critical steps, compared to the Diffusion-ES planner, the VLM-empowered system enables more flexible driving behaviors and complies with common sense in real-world driving. In these examples, the VLM assigns higher heuristic preference scores to favorable tactical decisions, allowing the selector to avoid errors produced by the rule-based scorer.}
  \label{fig:qualitative}
\end{figure*}

\Cref{tab:interplan_breakdown} compares the same 335 InterPlan scenarios. SAH-Drive (switch) scores 78.30 R-CLS/94.63\% R-SR, versus 71.15/90.15\% and 70.95/85.07\% for its fixed learning and PDM outputs. The 7.15--7.36 R-CLS-point gain supports planner complementarity. The fixed learning output excels at progress (98.21\%) and passing stopped vehicles (86.67\%); fixed PDM better respects the drivable area (98.21\% versus 92.24\%). The switch retains near-learning progress and passing scores (97.91\% and 83.33\%) while lifting drivable-area compliance to 98.81\%, consistent with adaptive use of both branches' strengths. PRIMA$^{\ddag}$ scores close to the fixed outputs overall and leads in no-at-fault-collision compliance (98.81\%), drivable-area compliance (99.40\%), and goal-lane changes (76.24\%). Its remaining gap centers on progress and obstacle-passing multipliers. Fixed-output differences also reflect proposal generation, model training, and scoring parameters: SAH-Drive (PDM) exceeds independently run PDM-Closed (70.95 versus 69.67 R-CLS). Thus, their comparison does not isolate VLM reasoning.

\subsection{Qualitative Studies}\label{subsec: quali}

A qualitative comparison of our approach and Diffusion-ES can be witnessed in Fig.~\ref{fig:qualitative}, where two cases are extracted with the scenario type of \textit{changing lane} and \textit{waiting for pedestrian to cross}. 

\noindent \textbf{Scenario I. \textit{Changing Lane}:} In the first scenario, as a pedestrian suddenly appears on the lane, the Diffusion-ES planner applies a braking maneuver to avoid the collision. However, despite the pedestrian remaining stationary ahead, the ego vehicle fails to attempt to move around the object to continue driving. In contrast, our approach deals with the potential risk by changing to the right clear lane, similar to the expert trajectory. The VLM holds the highest confidence in accelerating and continuing the lane change as it finds that the strategy is both safe and efficient.

\noindent \textbf{Scenario II. \textit{Waiting for Pedestrian to Cross}:} At the beginning of the second scenario, the ego vehicle is decelerating while approaching a junction to allow a nearby pedestrian to cross the street. The snapshots at the critical step indicate that as the pedestrian walks away, the Diffusion-ES planner gets trapped in a deadlock, unable to speed up due to the invalid scoring mechanism. The scene semantics is effectively captured by the VLM decision-maker, which expresses a high confidence to accelerate upon reasoning over the spatial relation between the pedestrian and ego vehicle.

\subsection{Ablation Studies}
We conduct ablation studies on Test14-Random benchmark to verify the effectiveness of our main components. \Cref{tab:ablation reason} shows the results of the ablation study on the reasoning process of our VLM-based decision-maker, while \Cref{tab:ablation score} presents the effectiveness of each component in the preference-informed trajectory selection in Eq.~\eqref{eq: conf-aware select}. In both tables, the configurations and results corresponding to our complete model are presented in \colorbox{newlightgreen}{olive green}, and the best performance on each metric is marked in \textbf{bold}.

\noindent \textbf{Ablation Study on the Reasoning Process of the VLM-Based Decision-Maker:} We refer to our original VLM-based decision-maker as \underline{S.U. \(+\) Two-Stage Top-K Dec.} Two variants of our original method are investigated:
\begin{itemize}
	\item \underline{S.U. \(+\) Single Dec.}: The VLM is instructed to perform scene understanding and then choose a single decision from the action space, which is similar to many existing pipelines~\cite{wendilu, zheng2024planagent}.
	\item \underline{S.U. \(+\) One-Stage Top-K Dec.}: The VLM is instructed to perform scene understanding and then choose Top-K decisions from the action space along with confidences.
\end{itemize}

Results in \Cref{tab:ablation reason} show that all modified reasoning designs underperform the original one across all metrics. The \underline{S.U. + Single Dec.} variant performs worst, even falling short of Diffusion-ES (\Cref{tab:planner compare}), because it omits tactical alternatives and reduces the system to a single decision-then-planning pipeline. The \underline{S.U. + One-Stage Top-K Dec.} variant is closer to the full model, suggesting that separating candidate generation from preference assessment provides an empirically useful selection signal.

\begin{table}[t]
    \centering
    \caption{Results of ablation study on the reasoning process of the VLM-based decision-maker on Test14-Random.}
    \resizebox{0.9\linewidth}{!}{
    \begin{tabular}{l|cccc}
        \toprule
        \textbf{Reasoning Process} & \textbf{NR-CLS}  & \textbf{NR-SR(\%)}  & \textbf{R-CLS}  & \textbf{R-SR(\%)} \\ 
        \midrule
         S.U. \(+\) Single Dec. & 81.35 & 91.57 & 82.42 & 93.49 \\ 
         S.U. \(+\) One-Stage Top-K Dec. & 85.97 & 94.25 & 86.19 & 94.64 \\
         \rowcolor{newlightgreen}
         S.U. \(+\) Two-Stage Top-K Dec. & \textbf{86.70} & \textbf{95.02} & \textbf{87.11} & \textbf{95.79} \\
        \bottomrule
    \end{tabular}
    }
    \label{tab:ablation reason}
\vspace{-0.5em}
\end{table}

\begin{table}[t]
    \centering
    \caption{Results of ablation study on the preference-informed trajectory selection on Test14-Random.}
    \resizebox{0.9\linewidth}{!}{
    \begin{tabular}{ccc|cccc}
        \toprule
        \((c^k_t)^{\omega_c}\) & \((J^k_f)^{\tilde{\omega}_f}\) & \((J^k_g)^{\tilde{\omega}_g}\) & \textbf{NR-CLS}  & \textbf{NR-SR(\%)}  & \textbf{R-CLS}  & \textbf{R-SR(\%)} \\ 
        \midrule
         \checkmark & & & 82.57 & 93.10 & 82.84 & 93.49 \\ 
         & \checkmark & & 74.02 & 80.84 & 73.71 & 80.08 \\
         & & \checkmark & 85.31 & 93.87 & 84.97 & 93.49 \\
         \checkmark & \checkmark & & 83.05 & 93.49 & 83.13 & 95.02 \\ 
         \checkmark & & \checkmark & \textbf{86.70} & 94.64 & 85.77 & 93.87 \\
         & \checkmark & \checkmark & 85.52 & 93.87 & 85.24 & 93.10 \\
         \rowcolor{newlightgreen}
         \checkmark & \checkmark & \checkmark & \textbf{86.70} & \textbf{95.02} & \textbf{87.11} & \textbf{95.79} \\
        \bottomrule
    \end{tabular}
    }
    \label{tab:ablation score}
\vspace{-0.5em}
\end{table}

\begin{figure*}
  \centering
    \includegraphics[width=0.8\linewidth]{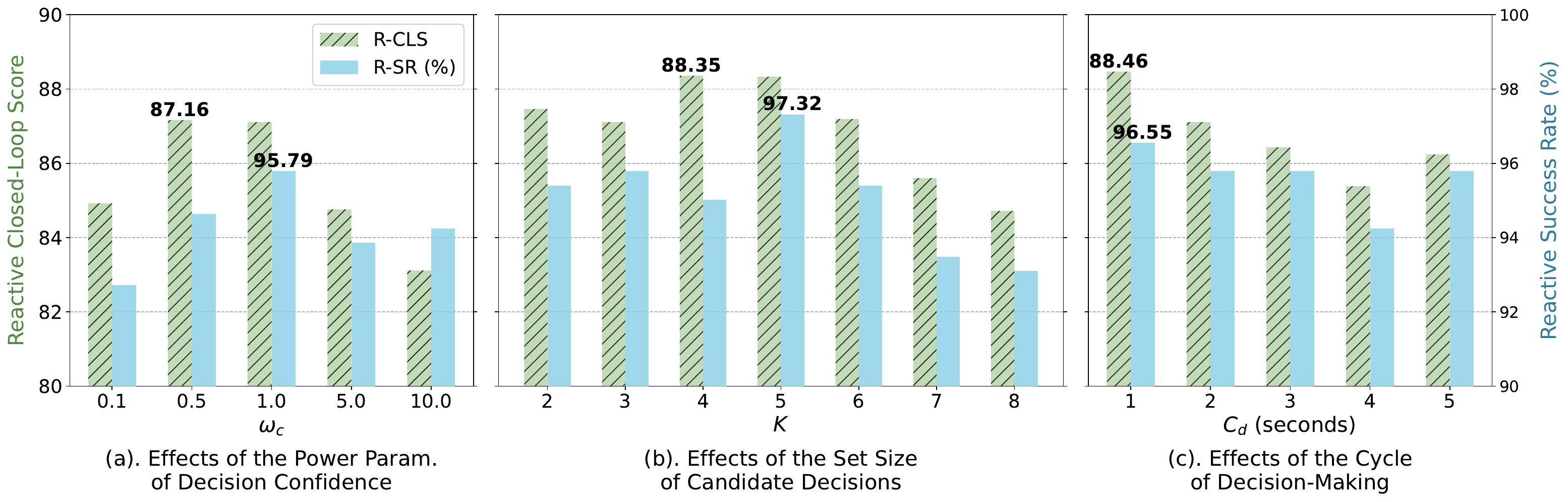}
    \vspace{-0.25cm}
  \caption{Quantitative effects of the key parameters in our architecture evaluated on Test14-Random under the setting of closed-loop reactive test. The best performance on each metric is annotated with text in \textbf{bold}.}
  \label{fig:param effects}
  \vspace{-0.5em}
\end{figure*}

\noindent \textbf{Ablation Study on the Preference-Informed Trajectory Selection:}
Our original trajectory scorer Eq.~\eqref{eq: conf-aware select} combines the heuristic preference $c^k_t$, decision-following score $J^k_f$, and general trajectory score $J^k_g$. \Cref{tab:ablation score} shows that the full combination obtains the best point estimates. Using only $c^k_t$ reduces NR-CLS by 4.13 and R-CLS by 4.27, demonstrating that the verbalized signal cannot substitute for explicit planning-quality evaluation. Combining $c^k_t$ and $J^k_g$ performs similarly to the full scorer in non-reactive tests but lags in reactive scenarios, while using only $J^k_f$ is substantially worse. These results support treating confidence as one heuristic input within a composite scorer.

\subsection{Analysis of the Impact of Key Parameters}\label{subsec: para effect}
The choice of key parameters directly affect the proposed method’s driving performance. We investigate three key parameters: decision confidence power \(\omega_c\), candidate decision set size \(K\), and decision cycle \(C_d\), with quantitative effects on Test14-Random R visualized in Fig.~\ref{fig:param effects}.

\noindent \textbf{Effects of the Power Parameter of Decision Confidence:}
The power parameter \(\omega_c\) balances decision confidence and motion planning quality in the trajectory scorer.
Besides the original \(w_c=1\), we test 0.1, 0.5 (prioritizing planning) and 5.0, 10.0 (emphasizing decision-making).
As Fig.~\ref{fig:param effects} shows, optimal performance occurs at the original setting; performance declines with deviations in either direction.
At \(\omega_c=0.1\), R-SR falls to 92.72\%, near the baseline without decision confidence (93.10\%) in \Cref{tab:ablation score}, as decision confidence is suppressed. Conversely, \(w_c=\)~10.0 reduces R-CLS to 83.11, close to the baseline relying only on decision confidence (82.57) in \Cref{tab:ablation score} due to insufficient planning evaluation. Thus, balancing decision and planning importance yields optimal overall performance.

\noindent \textbf{Effects of the Set Size of Candidate Decisions:} The set size \(K\) denotes the number of candidate decisions inferred by the VLM-based decision-maker. We test \(K\) from 2 to 8. As shown in Fig.~\ref{fig:param effects}, the overall performance peaks at \(K\) = 5 (R-CLS: 88.32, R-SR: 97.32\%), with the R-SR surpassing that of PDM-Closed (96.55\% in \Cref{tab:planner compare}). However, performance monotonically decreases when \(K>\)~5, as excessive candidate strategies are scenario-inapplicable, complicate VLM reasoning, and induce hallucinations. Adaptive \(K\) adjustment based on traffic context may further improve performance.

\noindent \textbf{Effects of the Cycle of Decision-Making:} Our original decision cycle \(C_d\) is 2\,s. Fig.~\ref{fig:param effects} shows performance under varied \(C_d\). When \(C_d\) decreases to 1\,s, more frequent VLM responses to traffic changes improve performance (R-CLS: 88.46, R-SR: 96.55\%) versus the original setting. Larger \(C_d\) than 2\,s degrades both metrics. Future work should calibrate \(C_d\) to balance performance and computational costs.

\begin{table}
    \centering
    \caption{Paired closed-loop intervention on 50 confidence--planning-score conflict scenarios. $\Delta$ is preference-informed (PI) minus planning-score-only (plan.).}
    \label{tab:paired confidence intervention}
    \resizebox{\columnwidth}{!}{
    \begin{tabular}{l|r|rrr|rrr}
        \toprule
        \textbf{Subset} & \textbf{$N$} & \textbf{PI CLS} & \textbf{Plan. CLS} & \textbf{$\Delta$CLS} & \textbf{PI SR} & \textbf{Plan. SR} & \textbf{$\Delta$SR}\\
        \midrule
        All      & 50 & 81.34 & 76.49 & +4.85 & 90\,\% & 84\,\% & +6\,pp. \\
        Hard-R   & 25 & 70.16 & 64.69 & +5.47 & 80\,\% & 72\,\% & +8\,pp.\\
        Random-R & 25 & 92.52 & 88.29 & +4.24 & 100\,\% & 96\,\% & +4\,pp.\\
        \bottomrule
    \end{tabular}}
    \vspace{-0.5em}
\end{table}

{
\subsection{Empirical Analysis of Verbalized Confidence}\label{sec:confidence analysis}
We conduct a paired first-decision intervention that directly evaluates incremental decision utility when confidence and the planning score disagree. From the reactive Test14-Hard and Test14-Random logs, we take only the initial decision state of each scenario and identify states where (i) the preference-informed fusion and the planning-score-only selector choose different actions, (ii) the fusion-selected confidence is at least 0.8, and (iii) that action has a lower planning score than the planning-only choice. Without inspecting either counterfactual outcome, we deterministically select 25 scenarios from each benchmark by sorting SHA-256 hashes of benchmark--scenario tokens. This yields a balanced, reproducible set of 50 conflict states. For each identical initial state, one branch executes the preference-informed (PI) action for the first 2-s decision cycle and the other executes the planning-score-only (Plan.) action; both then use the same Diffusion-ES continuation and simulation settings.

\begin{figure}
\centering
\includegraphics[width=\linewidth]{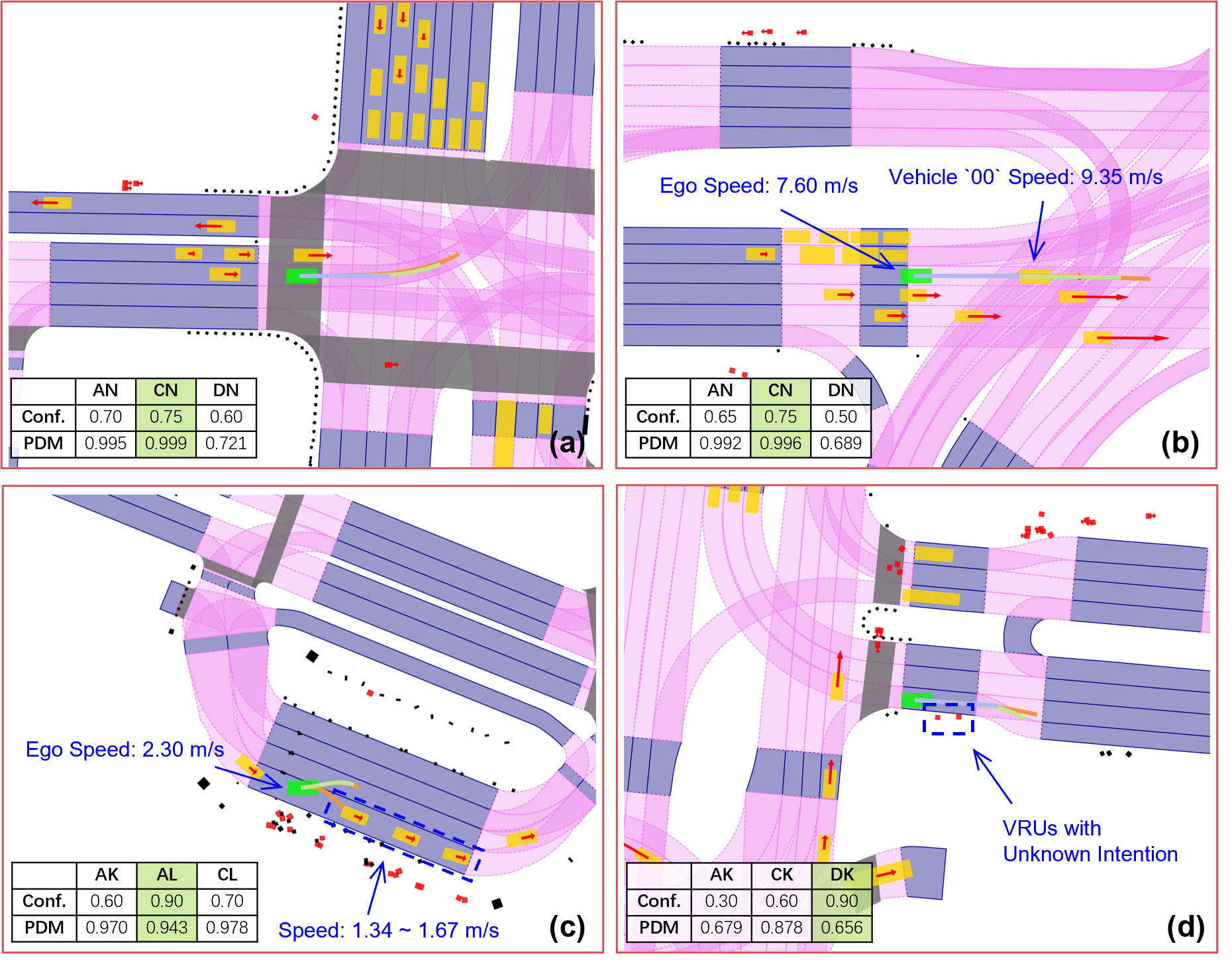}
\caption{Illustration of verbalized confidence and PDM scores in multimodal planning. Longitudinal actions are color-coded as \textcolor{fancyorange}{accelerating}, \textcolor{fancygreen}{cruising}, and \textcolor{fancyblue}{decelerating}. Snapshots (a) and (b) showcase the consistency between decision confidence and trajectory quality in moderate-confidence scenarios. The VLM favors cruising to balance safety, efficiency and comfort in clear traffic or when following a faster vehicle, resulting in the highest PDM scores. Snapshots (c) and (d) depict high-confidence scenarios that exhibit trade-offs. Despite the lowest PDM scores, in (c), the VLM prioritizes lane-change to detour around multiple slower vehicles ahead for long-term efficiency; in (d), it opts for decelerating to ensure safety near VRUs with unknown intention.}
\label{fig:multimodal vis}
\vspace{-0.5em}
\end{figure}

As shown in \Cref{tab:paired confidence intervention}, preference-informed selection improves the overall point estimates by 4.85 CLS and 6 percentage points in SR. The success contingency contains 42 joint successes, three successes unique to preference-informed selection, no success unique to planning-only selection, and five joint failures. The rescued scenarios directly demonstrate cases in which a \textbf{high-confidence/lower-planning-score tactical decision improves the final closed-loop outcome.} We therefore interpret this experiment as directional evidence of incremental long-term decision utility. Together with the full-policy ablation in \Cref{tab:ablation score}, it supports the use of verbalized confidence as an empirical preference signal for candidate arbitration while retaining explicit planning-quality terms. In Fig.~\ref{fig:multimodal vis}, we further illustrate representative cases showing consistency between the confidence and PDM metric in moderate-confidence scenarios and trade-offs in high-confidence scenarios, respectively. 
}

\begin{table}[t]
    \caption{Latency of foundation model inference on Test14-Hard.}
    \centering
    \resizebox{0.9\columnwidth}{!}{%
    \begin{tabular}{l|ccc}
    \toprule
    \textbf{Model} & \textbf{Inference Time} & \textbf{NR-CLS} & \textbf{R-CLS}\\
    \midrule
    PlanAgent$^{*}$~\cite{zheng2024planagent} & 5568\,ms & 72.51 & 76.82\\
    DualAD (GLM-4-Flash)~\cite{wang2025dualad} & 43\,ms & 63.38 & 66.40\\
    DualAD (GPT-4o)~\cite{wang2025dualad} & 2.7\,ms & 60.87 & 64.57\\
    PRIMA & 4747\,ms & 77.39 & 78.13\\
    PRIMA$^{\ddag}$ & 4793\,ms & 75.03 & 78.46\\
    PRIMA-Lite & 147\,ms & 67.89 & 77.26\\
    \bottomrule
    \end{tabular}}
    \label{tab: runtime}
\end{table}

\subsection{Analysis of the Runtime of Foundation Model Inference}
The main computational bottleneck of our original framework stems from the VLM-based decision inference. As shown in \Cref{tab: runtime}, PRIMA and its memory variant achieve average inference times of 4747\,ms and 4793\,ms, respectively, comparable to PlanAgent's 5568\,ms latency. This $\sim$5\,s inference time is a common characteristic of LLM/VLM-empowered frameworks adopting structured CoT reasoning~\cite{wendilu}, as the autoregressive generation of multi-step reasoning content dominates the computational cost.
In contrast, DualAD achieves drastically lower latency by directly outputting reference speeds, while obtaining lower CLS on the evaluated long-tail benchmarks. This comparison highlights the need to assess decision-output structure and downstream planning alongside inference time.

The original PRIMA inference latency exceeds the 2\,s decision cycle, precluding strict online deployment. PRIMA-Lite provides a lower-latency variant by predicting three tactical labels directly. On Test14-Hard, its average decision-inference times are 152\,ms (NR) and 141\,ms (R), averaging 147\,ms across the two settings. Planning comprises candidate preparation and shortlisting, trajectory rollout, and trajectory scoring; its total time is 242\,ms in NR and 293\,ms in R. The resulting decision-plus-planning times are 394\,ms and 434\,ms, respectively, below 0.5\,s and the 2\,s decision cycle.

\subsection{Analysis of Failure Cases}
\begin{figure}[t]
\centering
\includegraphics[width=\linewidth]{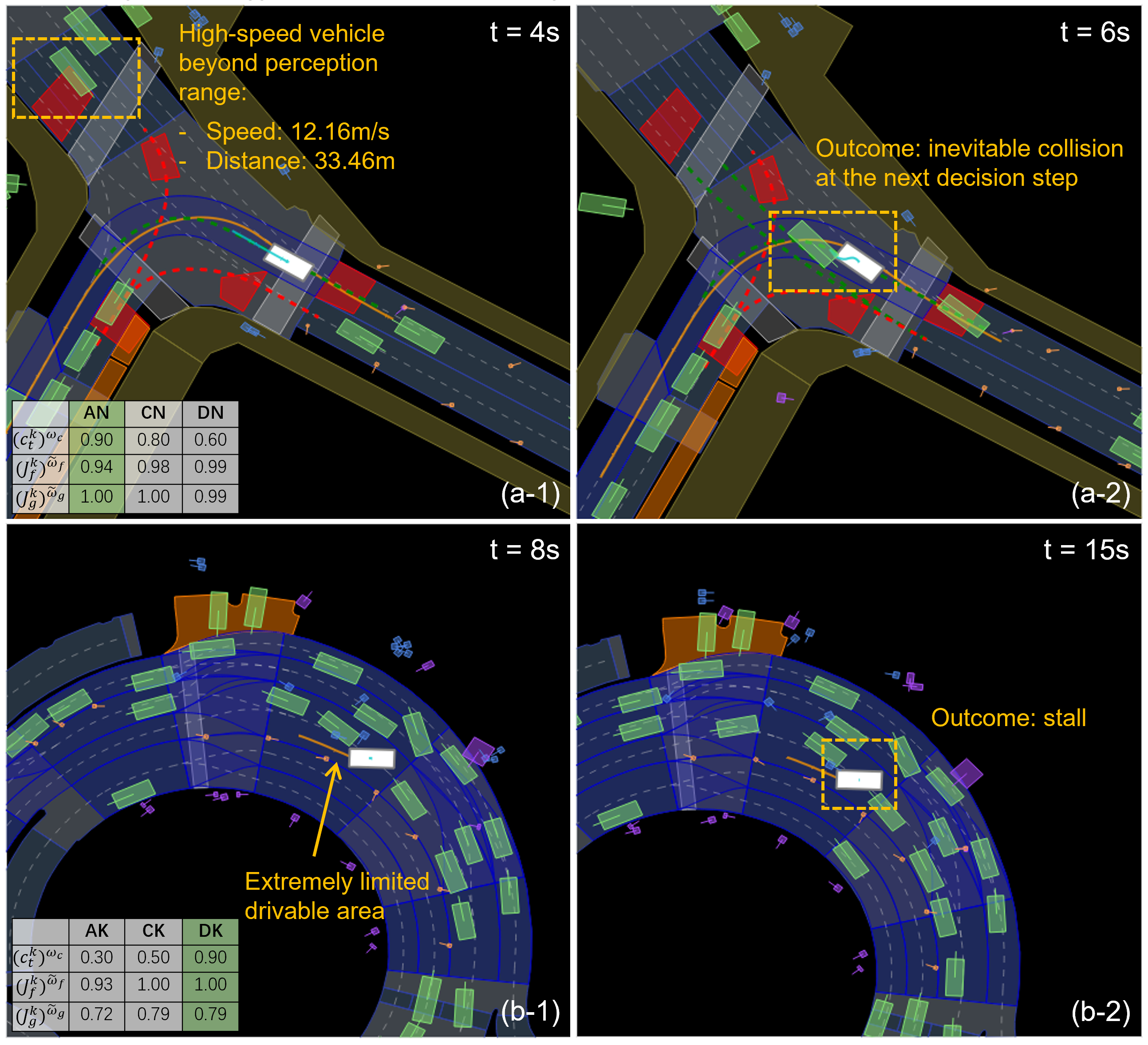}
\caption{Representative failure cases for analysis. (a-1) and (a-2): at-fault collision due to the insufficient perception range and fixed decision cycle; (b-1) and (b-2): low progress due to conservatism in the extremely limited drivable area.}
\label{fig:failure vis}
\end{figure}

We visualize two representative failure cases in Fig.~\ref{fig:failure vis}. 

\noindent \textbf{(a-1) and (a-2).} At $t=\,$4\,s, a high-speed oncoming vehicle is at a distance of 33.46\,m, beyond the predefined perception range (30\,m). This causes overconfidence in accelerating to complete the left-turn without awareness of the conflict. In the next decision cycle, the ego vehicle loses clearance to avoid the crash. This indicates the need for more rational selection of interactive objects in the scene, and warrants a more adaptive approach to determining decision intervention timing.

\noindent \textbf{(b-1) and (b-2).} At $t=\,$8\,s, the ego vehicle is confronted with a severely restricted drivable area due to an adjacent vehicle, a VRU, and a traffic cone. While the human expert swiftly maintains the movement, our VLM decision-maker is strongly biased on decelerating, with the downstream planner failing to yield a trajectory that allows for accelerating both safely and efficiently. The stall results in a 0 score on progress. This implies that our VLM decision-maker needs more experience accumulated on such edge cases to perfectly handle extreme hard instances, and that the planner should be further enhanced to better guarantee trajectory optimization quality.

\subsection{Performance Comparison Under Different Foundation Models}\label{subsec: lmm effect}
\begin{figure}[t]
\centering
\includegraphics[width=\linewidth, trim=0 50 0 40, clip]{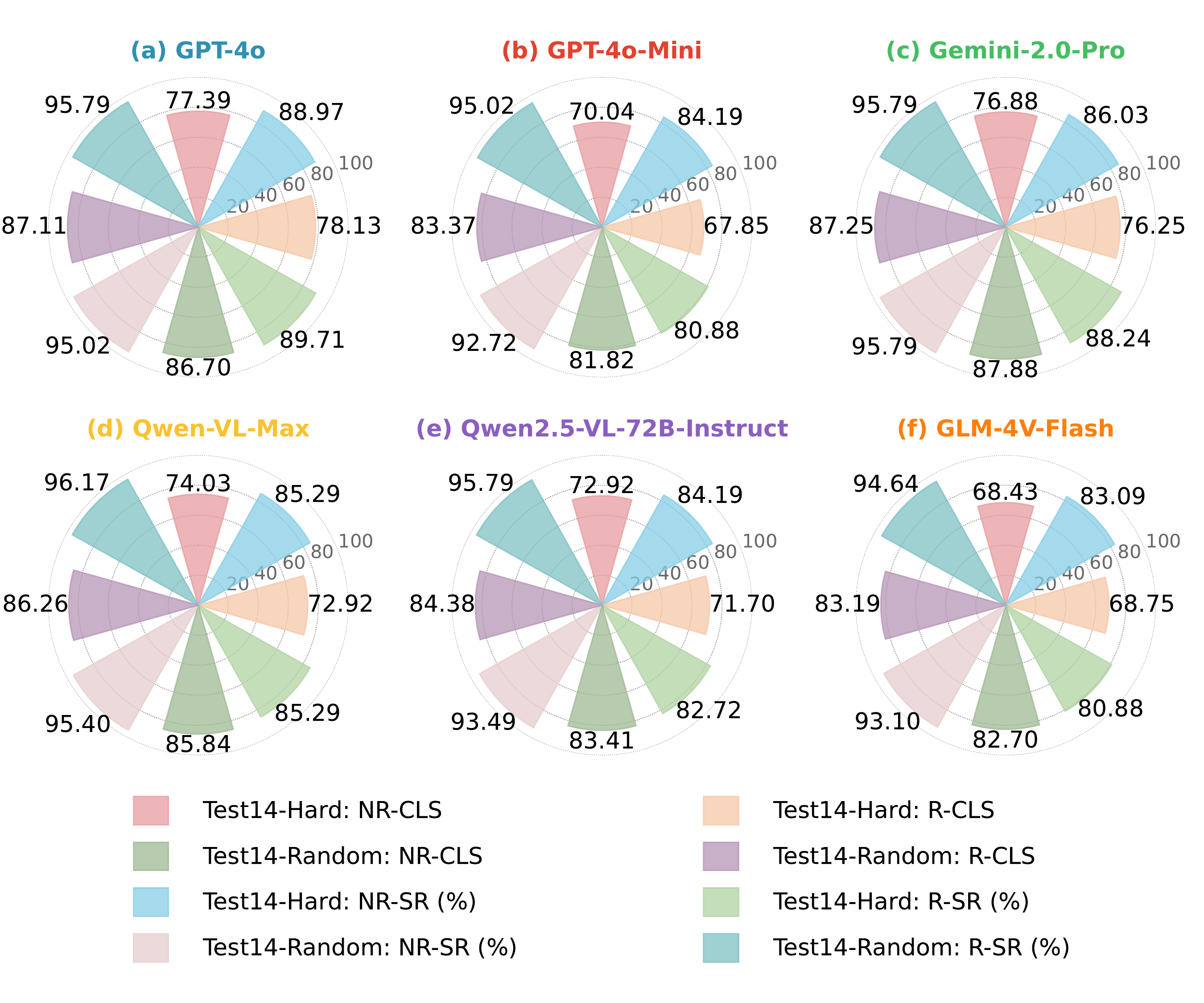}
\caption{Quantitative performance comparison of PRIMA powered by different foundation models on Test14-Hard and Test14-Random. The subplot titled with GPT-4o correspond to our original version.}
\label{fig:perform diff found}
\end{figure}

To explore the performance of PRIMA under different foundation models, we evaluate it using advanced VLMs (Gemini-2.0-Pro, Qwen-VL-Max, Qwen2.5-VL-72B-Instruct) and simplified ones (GPT-4o-Mini, GLM-4V-Flash). Fig.~\ref{fig:perform diff found} shows better performance with GPT-4o (best on Test14-Hard), Gemini-2.0-Pro (overall best on Test14-Random), and Qwen-VL-Max (highest R-SR=96.17\% on Test14-Random). Simplified VLMs perform worse than advanced ones, with larger gaps on Test14-Hard. 

To reveal how model characteristics affect the performance, Table~\ref{tab:diff lmm output} characterizes their outputs via \textit{Decisiveness} (mean Top-1/Top-2 confidence gap), \textit{Tactical Distribution Entropy (TDE)} (Shannon entropy of the frequency distribution of Top-3 action combinations), and \textit{Parse SR} (success rate of output formatting). Top-performing models GPT-4o and Gemini-2.0-Pro exhibit moderate decisiveness and relatively lower TDE, indicating balanced confidence and stable tactical preferences. Conversely, simplified models (\eg, GPT-4o-Mini) show excessive decisiveness (0.2582) and the highest TDE (1.8789), reflecting blind overconfidence and tactical dispersion. Qwen2.5-VL-72B-Instruct displays the lowest decisiveness (0.1558), indicating cognitive hesitation. Furthermore, the lower Parse SR of simplified models frequently degrades the pipeline into a decision-less fallback mode, compromising driving smoothness and consistency. These findings imply that advanced foundation models are still required  to ensure cognitive soundness and robust driving performance.

\begin{table}[t]
    \centering
    \caption{Quantitative characterization of decision-making outputs across different foundation models.}
    \resizebox{\columnwidth}{!}{
    \begin{tabular}{l|ccc}
        \toprule
        \textbf{Model} & \textbf{Decisiveness}  & \textbf{TDE}  & \textbf{Parse SR (\%)} \\ 
        \midrule
         GPT-4o                    & 0.1619  & 1.3073  & 100.00  \\ 
         GPT-4o-Mini               & 0.2582  & 1.8789  & 95.77  \\
         Gemini-2.0-Pro            & 0.1716  & 1.2948  & 100.00 \\
         Qwen-VL-Max               & 0.1569  & 1.6403  & 99.53  \\
         Qwen2.5-VL-72B-Instruct   & 0.1558  & 1.6398  & 99.26  \\
         GLM-4V-Flash              & 0.2144  & 1.7386  & 95.98 \\
        \bottomrule
    \end{tabular}
    }
    \label{tab:diff lmm output}
\end{table}

\section{Conclusion}
This study proposes PRIMA, a preference-informed, VLM-empowered autonomous driving framework. Our approach enables the VLM to consider multiple candidate decisions and verbalize confidence scores expressing their relative tactical preferences through a structured tri-stage reasoning process. By combining tactical preferences with motion-planning quality, the proposed framework achieves competitive and consistent performance across non-reactive and reactive closed-loop settings. 
PRIMA-Lite further shows that direct student-model decisions and static-candidate planning can substantially shorten inference latency while retaining useful closed-loop performance. 
In addition, the ablation studies conducted showcase the efficacy of different components in our architecture. 
A paired controlled intervention provides directional evidence that the signal contributes incremental long-term decision utility in confidence--planning-score conflicts.

Further improvements can be made to the current framework. First, a smooth conversion of the VLM's high-level decision to trajectory planning could be investigated, which would better benefit driving comfort especially during the switching of different maneuvers. Second, it is encouraging to consider the introduction of another agent to assess whether an embodied decision-making should be activated. It is expected to boost the potential of handling long-tail cases and simultaneously reserve the inference efficiency for normal cases. Third, building on PRIMA-Lite, future work could use planner-guided distillation with closed-loop feedback and targeted augmentation of rare maneuvers to improve the student's long-tail performance while retaining low inference latency. Finally, the reduced inference latency motivates hardware-in-the-loop and real-vehicle evaluations to verify robustness against perception noise under realistic operating conditions.

\bibliographystyle{ieeetr}
\bibliography{main}

\end{document}